\documentclass[acmsmall,anonymous=false,manuscript,nonacm]{acmart} 
\usepackage[utf8]{inputenc}
\usepackage{blindtext}
\usepackage{subfiles}   
\usepackage{graphicx}
\usepackage{array}
\usepackage{lscape}

\usepackage{natbib}

\copyrightyear{2024}
\setcopyright{rightsretained}

\title[Using Social Cues to Recognize Task Failures for HRI]{Using Social Cues to Recognize Task Failures for HRI: \\
Overview, State-of-the-Art, and Future Directions}
\author{Alexandra Bremers}
\affiliation{%
  \institution{Cornell Tech}
  \city{New York, NY}
  \country{USA}
}
\author{Alexandria Pabst}
\affiliation{%
  \institution{Accenture Labs}
  \city{San Francisco, CA}
  \country{USA}
}
\author{Maria Teresa Parreira}
\affiliation{%
  \institution{Cornell Tech}
  \city{New York, NY}
  \country{USA}
}
\author{Wendy Ju}
\affiliation{%
  \institution{Cornell Tech}
  \city{New York, NY}
  \country{USA}
}


\begin{document}
\begin{abstract}
 Robots that carry out tasks and interact in complex environments will inevitably commit errors. Error detection is thus an essential ability for robots to master to work efficiently and productively. People can leverage social feedback to get an indication of whether an action was successful or not. With advances in computing and artificial intelligence (AI), it is increasingly possible for robots to achieve a similar capability of collecting social feedback. In this work, we take this one step further and propose a framework for how social cues can be used as feedback signals to recognize task failures for human-robot interaction (HRI). Our proposed framework sets out a research agenda based on insights from the literature on behavioral science, human-robot interaction, and machine learning to focus on three areas: 1) social cues as feedback (from behavioral science), 2) recognizing task failures in robots (from HRI), and 3) approaches for autonomous detection of HRI task failures based on social cues (from machine learning). We propose a taxonomy of error detection based on self-awareness and social feedback. Finally, we provide recommendations for HRI researchers and practitioners interested in developing robots that detect task errors using human social cues. This article is intended for interdisciplinary HRI researchers and practitioners, where the third theme of our analysis provides more technical details aiming toward the practical implementation of these systems. 
 \end{abstract}

\maketitle

\section{Introduction}

Robots are increasingly deployed to work with and amongst people. This creates challenges for robots, but also presents an opportunity: 
robots can watch people's reactions to them to help recognize when they have committed an error. As \citet{honig2021expect} describe in their Theory of Graceful Extensibility, although it is impossible to eliminate unexpected robot failures completely, it is possible to design robots that adapt to newly emerging contingencies by leveraging the socio-technical human-robot ecosystem to repair failures and adapt to the environment. \citet{lewis2009using} advocate the idea of using \textit{humans as sensors} to use human behavior to inform a robot whether its actions were successful or unsuccessful. After all, a myriad of social cues--from subtle cues like eye gaze to more overt cues like language and gestures-- are used by people to communicate context, failure, and success to one another \cite{wang2022systematic, lin2020review}. 

\begin{figure}
\includegraphics[width=0.8\textwidth]{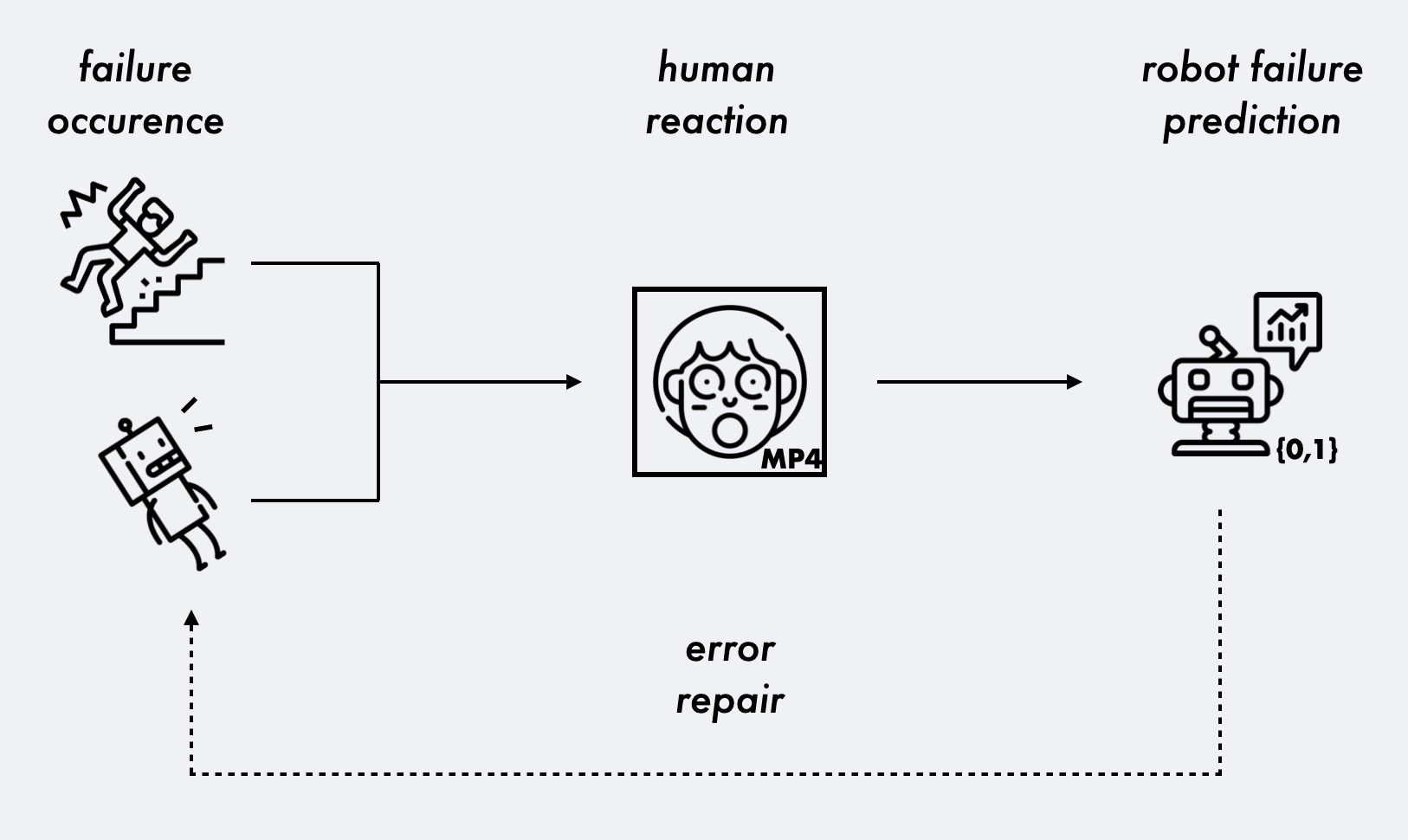}
\caption{Schematic overview of interaction intelligence for task failure detection through human social cues. A failure occurs (left), leading to a human reaction (center), which is used as data input for failure prediction from the robot (right). Detection of failure is an important first step for error repair in HRI.\protect\footnotemark}
\label{fig_schematicconcept}
\end{figure}

Since error recognition is a prerequisite to repair, human social cues can provide additional cues for error detection that generalize across conditions, and these can be used in conjunction with task-specific models for robotic error detection to improve robot performance. This approach, which we illustrate in \autoref{fig_schematicconcept}, is gaining traction in human-robot interaction and include \citet{stiber2022effective, stiber2023using, 2020stiber, 2022stiber, cuadra2021look, cuadra2021my, bremers2023bystander,2020kontogiorgosembodiment,kontogiorgos2020behavioural, kontogiorgos2021systematic, 2016hayes,2017trung,2019morales,2010aronson,honig2021expect,parreira2024study,candon2023nonverbal,candon2023nonverbalhuman,candon2024leveraging,zhang2023self} and others. 

\subsection{Contribution statement}
In this current work, we seek to contextualize recent HRI efforts within a larger body of work on social cues as feedback to map out a broader research agenda on social cue recognition for human-robot interaction. Key benefits of social cue recognition include the increased availability of feedback, the potential for more natural interactions, and the fundamental inclusion of the human interactant as a stakeholder in determining task success. 

\footnotetext{Icon sources: Freepik, Smashicons, Chattapat, wanicon via Flaticon.com.}

We present a framework and research agenda to build upon existing literature reviews on failure, such as those of \citet{honig2018understanding}, \citet{giuliani2015systematic}, and \citet{tian2021taxonomy}. Whereas this prior work seeks to deepen our understanding of failure in the context of HRI, our work focuses on tangible methods to test and implement social responses to failures as a way of improving human-robot interaction. This work also builds upon machine learning-focused work on reinforcement learning for HRI and general human activity recognition techniques \cite{akalin2021reinforcement,kulsoom2022review}. By bringing together disparate lines of related work from computer science, behavioral science, and human-robot interaction, we can better address how social cues can be used to identify task failures.

\subsection{Scope}
Our aim in this work is to set out a framework and research agenda based on an understanding of the state-of-the-art in the following three research areas:
\begin{enumerate}
\item How is feedback from human social cues defined in \textbf{behavioral science}?
\item How are task failures recognized in \textbf{Human-Robot Interaction (HRI)} research?
\item What are state-of-the-art \textbf{machine learning tools and datasets} for failure detection in HRI with social cues?
\end{enumerate}
This article thus unites insights from behavioral science, human-robot interaction, and machine learning. Importantly, we focus on error \textit{detection} and do not go in-depth into error \textit{repair} strategies.

\subsection{Approach}

We gathered key papers from HRI topics and ML topics that investigated task failures between humans and robots and imported those papers into the Citation Gecko software tool \cite{walker_2022} to generate connected papers that cited or have been cited by several seed papers, as well as conducted searches through databases containing academic literature like Association for Computing Machinery (ACM) Digital Library and Google Scholar. The papers included in this article are the papers that we subjectively believe are both the most relevant and likely to have the most potential impact in the field. This work does not aim to achieve a replicable review of all papers in the field. Instead, by sharing this work, we strive to start a discussion and lay the groundwork for future research in robot error detection based on human social cues. We delve into these three themes separately in the main body of the paper and explore recent literature reflecting the field's current state. An essential contribution of our work is that it summarizes the state-of-the-art use of machine learning solutions to recognize or infer task failures for people, robots, and machines -- providing a starting point for practical applications. 
\section{Theme 1: Human social cues as feedback: perspectives from behavioral science}
\label{sec:theme1}

\citet{fiore2013toward} define social cues as "biologically and physically determined features salient to observers because of their potential as channels of useful information." While there exist many types of definitions and taxonomies of social cues and signals, we take on this working definition as it has been accepted for use in human-computer interaction (HCI), and will aid our analysis of machine learning approaches to applying social cues in HRI.
The widely accepted Computers Are Social Actors (CASA) paradigm states that people interpret computer behavior as social cues and express social cues towards computers \cite{nass1994computers}. One can assume that social cues that people direct at a computer will be similar to those that people direct at another person. \citet{leathers1976nonverbal} proposed a taxonomy of interpersonal communication types: verbal, visual, auditory, and invisible. \citet{feine2019taxonomy} built on this work to propose a taxonomy of social cues in conversational agents. 

To understand how human social cues could be used as input for robots to detect failures, we discuss existing behavioral science literature to address the following questions:
\begin{itemize}
    \item What is human error?
    \item How do people behave when they detect another person's mistake?
\end{itemize}

We introduce a working taxonomy of social signal-based error detection depicted in \autoref{tab:errortax}. A \textbf{Type 3} error recognition method, where there is no self-awareness about the error but can be recognized by others, could aid robots to become better at the detection of errors previously unknown. We intentionally kept this taxonomy simple -- it is not meant to be comprehensive at characterizing all types of errors. Rather, it serves as a tool for researchers and practitioners to consider errors from a perspective of \textit{social awareness}.

\begin{table}[h]
\caption{To help conceptualize approaches to error detection, we introduce a taxonomy of error detection based on self-awareness and social feedback.}
\label{tab:errortax}
\begin{tabular}{c|c|c|l}
\textbf{Type} & \textbf{Self} & \textbf{Other(s)} & \textbf{Description}                                        \\ \hline
1             & N             & N                 & "I don't recognize an error, and neither does anyone else." \\
2             & Y             & N                 & "I recognize the error, but nobody else does."              \\
3             & N             & Y                 & "I don't recognize the error, but others do."               \\
4             & Y             & Y                 & "I recognize the error, and so do others."                 
\end{tabular}
\newline
\end{table}

\subsection{What is human error?}
\label{sec:3-1-humanerror}

A key issue in human error research is the lack of a unified definition of human error \cite{read2021state}. The terminology around the topic of error can include various words that are, at times, used interchangeably, such as failures, slips, mistakes, errors, and unintentional actions. \citet{hollnagel1991phenotype} defines errors as "actions not as planned" and describes two ways in which actions can fail: mistakes (the plan was incorrect) and slips (the execution of a correct plan was incorrect). Furthermore,  errors have a genotype (the functional aspects that contributed to the error) and phenotype (how the error appears) \cite{hollnagel1991phenotype}. Within definitions of human error, errors are divided into categories through various taxonomies. These exist along all sorts of dimensions, including social vs. technical errors \cite{honig2022taxonomy}, benign vs. catastrophic errors \cite{laprie1985dependable}, recoverable errors or less so \cite{o2004demonstrating}, and so on. In our work, we will define errors as "actions not as intended'' -- where the ultimate judgment of the action's success lies with the stakeholder(s) of the interaction. Examples of such actions could thus include errors like slipping on a banana peel, farting in public, or walking around with a ketchup stain on one's shirt. What matters is the stakeholder's reaction and judgment. 

Research on human errors is extensive. Much of this work relates to task analysis -- techniques to understand users' tasks. Task analysis has been applied in several domains, including piloting \cite{rouse1983analysis} and industrial installation \cite{rasmussen1982human}, and more recently in healthcare applications \cite{lane2006applying, phipps2008human}. \textit{Error} analysis adds a focus specifically on tasks that do not succeed. \citet{read2021state} review the key perspectives, theories, and methods in human error research from the 1960s until now from the lens of human factors. They describe a shift from looking at human-technology interaction to viewing errors in the context of socio-technical systems -- a perspective reminiscent of the field of complexity science. Here, errors are analyzed not as the behavior of an individual component but rather as a failing interaction between components or a system failure. 

The transition from technology-centered error analysis to a systems perspective of errors is akin to actor-network theory, which is a widely applied theory first described by the likes of Latour and Callon \cite{latour1996actor, callon1986sociology}, that states that actions and their components only exist in concert with one another and that a separation between social and technical relations is impossible \cite{tatnall2005actor}. Analogously, \citet{latour2007reassembling} states that a failure can only exist in the breakdown of interactions between its components. Systems only perform a function; what makes the outcome of the function successful is not inherently defined, but defined from the perspective of a human stakeholder. This view on successes and failures inherently emphasizes the ultimate human definition of whether or not an action is successful. In line with this complex view of the interactions between people and objects, users are not to be seen as only those who interact explicitly with a specific system. Within human-technology interaction research, voices have been calling for concern for users who, instead of intentionally interacting with technology out of intrinsic motivation, are forced to interact with technology. \citet{marsden1996human} describe this type of user as the \textit{accidental user}. These perspectives are important to take into account in designing human-robot interactions in complex environments, as the robot needs not only to be able to function directly with the operator but also with third parties in the space.

\subsubsection{Detecting errors}

\begin{figure}[t]
\makebox[0.5\textwidth][c]{\includegraphics[width=0.9\textwidth]{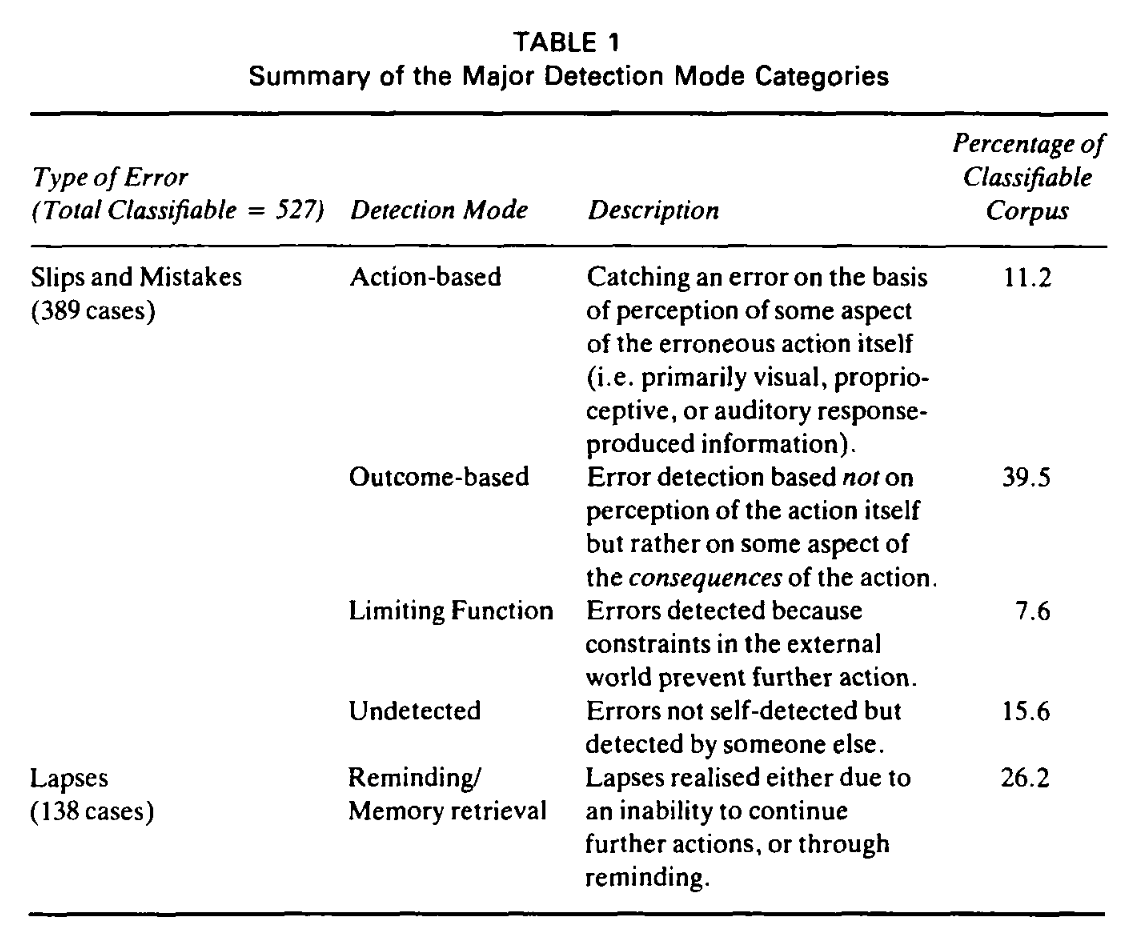}}
\caption{A human-robot failure taxonomy (reprinted with permission from \citet{sellen1994detection}, Table 1, \copyright Wiley).}
\label{fig_sellen_table1}
\end{figure}

Now how do we go from knowing what errors are to detecting them? In the paper "Detection of Everyday Errors," \citet{sellen1994detection} states that, despite the prevalence of literature describing errors, the \textit{detection} of errors had been an underdeveloped area. \textit{Error detection} refers to the awareness that an error has occurred, whereas \textit{error identification} concerns what has gone wrong and what should have happened, and \textit{error recovery} focuses on how to undo the error. Based on the analysis of 600 everyday slips and errors in a diary study, a theoretical taxonomy of (self) error detection methods was proposed. 
The errors were self-produced in everyday tasks -- that is, through the experimental setup --there was no staging of errors, nor was the focus on detecting errors by other people. The resulting descriptive taxonomy from \citet{sellen1994detection} is reproduced in \autoref{fig_sellen_table1} and covers the following categories of self-detection of errors: \textit{action-based} (perceiving the erroneous action), \textit{outcome-based} (perceiving the consequences of the action), \textit{limiting function} (perceiving external constraints preventing further action), \textit{undetected} (not self-detected, but detected by someone else) and \textit{reminding/memory retrieval}. 

We assume an analogous process for when people detect errors or failures of other people. However, the sensory input to these modes will be limited to what is observable about the other person (either the interactant in dyadic interactions, or the bystander). For instance, the primary modes of inferring that a person makes an error will likely involve visual and auditory information, with other senses in a more limited manner (e.g., without looking, one might be able to tell that another person has left the fridge door open, through hearing a change in sound, or even feeling the cold air escaping). 

Automated human activity recognition, as well as intention reading, may also inform the field of error detection through human behavior. \citet{bonchek2014towards} developed a model of intention recognition, specifically focusing on sequences of observed actions and the rationality of movements. Here, intention \textit{detection} means discerning whether a sequence of actions was intentional or without underlying intention. Intention \textit{prediction} is trying to extrapolate from a sequence of actions the likely end goal. Both use human behavioral cues, such as body position or gaze, to better understand the environment. \citet{epstein2020oops} presented the Oops! dataset of unintentional actions, which mostly includes human actions. Models for automated detection of action intention have been developed from this dataset \citet{epstein2021learning}. These and other examples are discussed in detail in Section \ref{sec:theme3}.

\subsection{How do people behave when they detect another person's mistake?}
\label{subsec:humanreactions}

Communication can be seen as a process of building mutual understanding about a context or situation. Mutual understanding can be achieved through grounding. Grounding sequences are communicative processes in dyadic interactions consisting of three stages: 1) one actor performs an action, 2) an addressee provides verbal or nonverbal feedback to signal understanding or correctness, and 3) the first actor acknowledges this signaling \cite{1991clark,2016hayes}. Grounding sequences can occur in all types of interactions, including erroneous interactions; similar processes are at play in non-dyadic interactions. The concept of using human social feedback for error detection in HRI builds upon the concept of grounding, as the human interactant provides feedback to signal correctness.

What kind of feedback gets communicated when a mistake is made? A relevant psychological phenomenon to consider here is embarrassment. According to \citet{keltner1997embarrassment}, embarrassment, like shame, is a self-conscious emotion, but follows from the breaking of a convention rather than a moral mishap, and elicits amusement rather than sympathy from observers. Observers' reactions can be communicated verbally or non-verbally -- the non-verbal communication consisting of all communicative aspects except speech \cite{mandal2014nonverbal}. This includes a combination of gaze, facial expressions, nonverbal utterances, nodding, body position, and proxemics, among other gestures. Thus, nonverbal communication is a key component in conveying empathy \cite{Haase1972}. For instance, in the case of hurting oneself \cite{bavelas1986show}, people use facial expressions to communicate they understand how someone feels when they are physically hurt. 

On the other hand, people are so attuned to others' perceptions of them that even perceptions of being watched can lead to measurably different neurological responses during task performance, which plays a role in error processing \cite{park2014interdependent}. Observed bystander reactions can thus influence the behavior of the self. This influence can have an intended outcome (e.g., as a means of social control \cite{edinger1983nonverbal}) or unintended consequences (for instance, by contributing to embarrassment \cite{parrott1991embarrassment}). \citet{blair2003facial} reports that viewing the facial expressions of another person while performing an action can modulate the likelihood that the action will or won't be performed in the future, which is especially relevant for responding to failure.  \citet{edinger1983nonverbal} cover a few examples of other research papers that highlight the effects of positively interpreted behaviors on increased confidence and task performance. Some of these works describe specific, often intentional, nonverbal behaviors that result in positive feedback and reinforcement, such as smiling, positive head nods, and increased eye contact. \citet{wang2016nonverbal} give an example of the effectiveness of nonverbal behavior as a form of social reinforcement: teachers for second-language acquisition classes often use nonverbal behavior along with corrective feedback. This nonverbal behavior can consist of hand- and head movements, affective displays, kinetographs, and emblems. The most common nonverbal behaviors were nodding, shaking the head, and pointing at a person or artifact. 

One final note concerns the interpretability of social cues and their causality. While early research on facial expressions mostly interpreted these as reflecting emotions, more recent work states that facial expressions also co-occur as a side effect of actions that are regulating (such as adaptation to light), protective reflexes (such as sneezing), or aiding in homeostatic processes (such as yawning) ~\cite{fridlund1997facial}. To avoid a false concentration on the interpretation of facial expressions as a way in which emotions are transmitted from one person to the public, ~\citet{fridlund1997facial} advocate for the usage of the term "facial behavior", along with the terms "emitter" and "observer". A single facial expression, such as yawning, can thus depend on many factors for which the cause is unclear. An approach of correlation rather than attempting to attribute causality will be most fruitful for HRI applications -- as long as one is aware of the limitations of this stance.

\subsection{In summary}
Among many definitions of human error, a useful definition is provided by ~\citet{hollnagel1991phenotype}, according to whom errors are "actions not as planned", either because the plan was incorrect ("mistakes"), or because the execution of a correct plan was incorrect ("slips"). Analysis of human errors mainly originated from the field of task analysis, but is moving towards a more complex and holistic view of errors, where successes and failures are not tied to a task but rather defined from the perspective of a human stakeholder. This new view is described by ~\citet{read2021state} and is in line with actor-network theory ~\cite{tatnall2005actor}.

Studies focusing on \textit{detecting} human errors have been sparse until the end of the last century ~\cite{sellen1994detection}. ~\citet{sellen1994detection} introduces a descriptive taxonomy of self-detection of errors, where one category, "unidentified errors," specifically describes errors that aren't self-detected but detected by someone else. This is where the opportunity lies for using human reactions as input data in human-robot interaction, helping the robot detect its own failure -- what we describe as \textbf{Type 3} errors, as per our introduced taxonomy. 

Different reactions to errors can include eyebrow raises, head movements, gaze, and facial expressions. Many studies point to the fact that social reactions to a person's mistake are commonly observable by the person and can positively influence the person's performance, both intentionally and unintentionally ~\cite{edinger1983nonverbal,parrott1991embarrassment,wang2016human,ergul2021case}. ~\citet{jones2011behavioral} even states that the behavioral change resulting from social feedback resembles reinforcement learning. 

We advise HRI researchers to treat social cues as a signal that can be correlated, rather than trying to derive intrinsic meaning from the social cues themselves, as many other confounding variables can influence social cues than just the robot failure at hand (see \cite{fridlund1997facial} for more details). In the next section, we will examine how the field of human-robot interaction approaches task failures -- are there similarities to error recognition mechanisms in human-human interactions, and what are the gaps?

\section{Theme 2: Recognizing task failures in human-robot interaction research}
\label{sec:theme2}

In this section, we cover prior work around the following questions:

\begin{itemize}
\item What types of robot errors exist?
\item How can robots harness nonverbal human feedback?
\end{itemize}

This theme focuses on literature clarifying our understanding of robot error and robot error detection, as well as how robots could leverage human social feedback for error detection.

\begin{figure}[h]
\makebox[0.5\textwidth][c]{\includegraphics[width=0.9\textwidth]{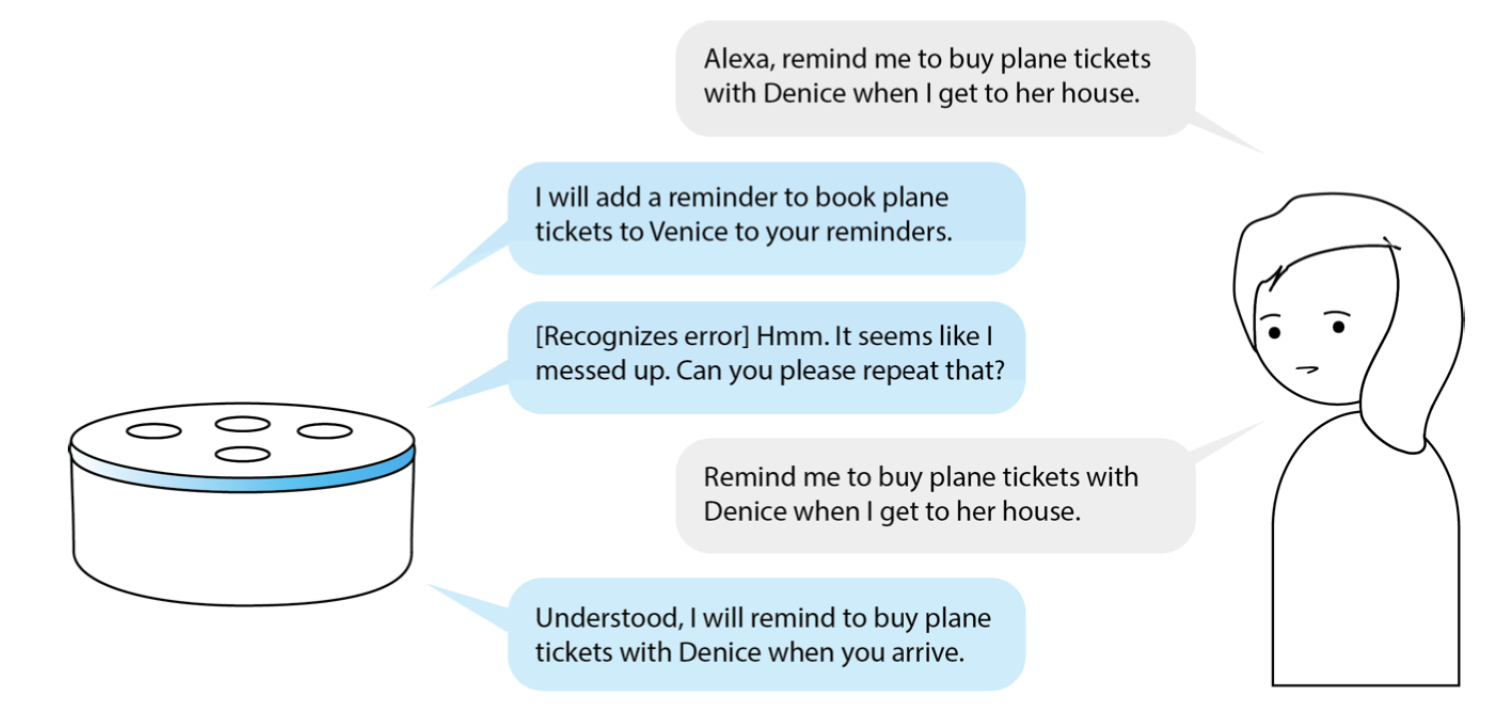}}
\caption{An illustrated example of the recognition of a conversational error and its subsequent repair that illustrates the complexity of errors (reprinted with permission from \citet{cuadra2021my}, Fig.1 \copyright ACM). We could imagine a similar scenario for a robot, where it recognizes an error based on social cues.}
\label{fig_cuadra2021fig1}
\end{figure}

\subsection{What types of robot errors exist?}

We can imagine a voice assistant that misunderstands a user command for something else (e.g., \citet{cuadra2021my}, \autoref{fig_cuadra2021fig1}). Looking at an example like this alerts us of the complexity and richness of errors committed by robots and non-robotic interactive agents. To illustrate the complexity of this space, robots can be either anthropomorphic or not, and perform tasks in similar ways to humans (e.g., how Alexa responds with spoken words) or tasks that have no human equivalent. Likewise, some robot errors can follow similar definitions to human errors, as we reviewed in Section \ref{sec:3-1-humanerror} (Theme 1), whereas for other errors these definitions will not hold (for example, hardware failures). 

\subsubsection{Taxonomies of robot failures}
\label{subsec}

\citet{tian2021taxonomy} proposed a taxonomy of social failures \textit{in HRI}, thereby expanding the prior taxonomy of social failures by \citet{honig2018understanding}. \citet{tian2021taxonomy} identify two main categories of robot failures: 1) social failures, resulting in lower perceived socio-technical performance, and 2) performance failures which are defined as technical exceptions in delivering a designed functional task. Social robot failures are further differentiated between \textit{interaction-} (with the environment, other agents, or humans) and \textit{technical} (hardware or software) failures \cite{tian2021taxonomy}. Each failure can be characterized by functional severity, social severity, relevance, frequency, condition, and symptoms. Based on this work, a new taxonomy of robot failures was recently suggested in \citet{honig2022taxonomy}, expanding the original taxonomy from \citet{honig2018understanding} to include both technical failures and the \textit{social} failures adopted from \citet{tian2021taxonomy}. 

Notably, there exist many other taxonomies of robot failures, such as taxonomies focusing on benign vs. catastrophic errors \cite{laprie1985dependable}, errors of varying recoverability \cite{o2004demonstrating}, physical vs. human errors \cite{carlson2005ugvs}, interaction vs. algorithm vs. software vs. hardware errors \cite{steinbauer2012survey}, and communication vs. processing errors \cite{brooks2017human}. \citet{nielsen2023using} used an ethnographic video analysis method on YouTube videos to code interaction breakdowns between humans and service robots and identified the failure categories of issues with input channels, detection failure, interaction breakdowns, and environmental disturbances. Due to the rich nature of the robot error, which involves many dimensions at once, an instance of a robot error can, similar to human errors (see Section \ref{sec:theme1} for Theme 1), be described along various dimensions from different taxonomies, informed by what is most useful for the audience and application at hand. This paper will mostly rely on the taxonomy derived from the iterative work by \citet{honig2018understanding,honig2022taxonomy}, and \citet{tian2021taxonomy}. This work is consistent with definitions in key related work on human reactions to robot failures, such as \citet{giuliani2015systematic}'s categorization of social signals during error situations.

\subsection{How can robots harness nonverbal human feedback?}
\label{subsec:harnesshri}

Robots using nonverbal behavior to impact human task performance -- the inverse of what we propose -- has been extensively studied \cite{bethel2007survey,mavridis2015review,Saunderson2019}. \citet{urakami2023nonverbal} provides an extensive theoretical overview along with proposed nonverbal codes for HRI from the perspective of communication studies. \citet{wallkotter2021explainable} reviews the explainability of embodied agents through a lens of social cues. Specific examples of studies where robots were equipped with social cues include \citet{cohen2017influence}, where researchers used a social humanoid robot (iCub) that provided facial feedback during a motor coordination task, finding that positive social feedback (iCub smiling) enhanced task performance. \citet{breazeal2005effects} studied the impact of a robot's nonverbal social cues, including gaze, nodding, and body motion, on collaborative task performance between a human and a robot, finding positive effects on understandability, task performance, and robustness to errors. Prior efforts have also looked into using the expressive qualities of non-anthropomorphic robot movement to communicate and collaborate with people \cite{takayama2011expressing,hoffman2014designing,mok2015performing,li2019communicating,bethel2007survey}.

Some HRI work has explored human reactions to failure. \citet{2019morales} provide a qualitative study of real-life human-robot interactions, where failures include bodily harm or property damage, and report on users' reactions and willingness to help. \citet{2019sanne} also explored the effect of the severity of robot errors on human-robot collaboration, finding that low-impact errors affected the trust in the robot's robustness for future collaborations. \citet{2020stiber} studied human reactions to varying degrees of robot error severity and found that humans respond faster to more severe errors and that responses to failure become multimodal as the error unfolds (e.g., from eyebrow raises to verbal utterances).
\citet{2015mirnig} carried a video analysis in a large corpus of humans interacting with failing robots and characterized human reactions, including reaction times and types of behavior displayed (verbal utterances and body motion, mostly). In these and other works, human reactions to robots failing have been found to be complex and include a great variety of behaviors, such as verbalizations \cite{kontogiorgos2021systematic}, body motion \cite{kontogiorgos2021systematic,2017trung,giuliani2015systematic}, gaze \cite{kontogiorgos2021systematic,2010aronson} and facial expressions \cite{kontogiorgos2021systematic,hwang2014reward,2020stiber}. 

While \textit{verbal} reactions to errors are insightful \cite{2005skantze,2013yasuda}, there is potential in leveraging nonverbal social cues. For example, robots can use computer vision to calculate a user's body position, which can be related to human affect \cite{2012mccoll}. This, in turn, can be used to evaluate aspects of the interaction. Some of this work has precedents in HRI. For instance, \citet{2011sanghvi} calculated the user's posture and body motion to compute children's engagement in a game with a social robot companion. Head and shoulder movement has also been used to detect robot failure \cite{2017trung}. Other sources of nonverbal behavior inputs are facial expressions \cite{2013cid} and eye gaze \cite{2014mehlmann}, among others. \citet{2016richter} used users' gaze and lip movement to improve the robustness of dialogue systems. \citet{2016huanggaze} also used human gaze in collaboratory HRI to anticipate human action. \citet{2016hayes} studied nonverbal behavior in reaction to robot failures in a learning-from-demonstration task. \citet{hwang2014reward} used the facial expressions of human observers, captured by a webcam, as input to a reinforcement learning system. \citet{broekens2007emotion} conducted a simulation study in which a robot (a simulated agent in a grid world) learned behavior through reinforcement learning.
      
While it falls outside the scope of this review, extensive work exists on error correction, both for human and robot errors. Examples of strategies employed in an HRI context are humor \cite{green2022s}, multi-robot collaboration \cite{reig2021flailing}, or narrative \cite{frederiksen2022robot}. These corrections can result in better interactions than when there was no error in the first place \cite{cuadra2021my}. For robots to correct errors, however, they first need to be able to \textit{identify} an error: leveraging human social cues provides new avenues to achieve this.  

\subsection{In summary}
Looking at individual errors (e.g., ~\citet{cuadra2021my}, Figure \ref{fig_cuadra2021fig1}) shows the complexity, richness, and diversity of HRI failures, and highlights the challenges in providing a generalizable taxonomy. We recommend the taxonomy presented by \citet{honig2022taxonomy} built based on the work of \citet{honig2018understanding} and \citet{tian2021taxonomy}. 

The field of HRI is starting to foster human nonverbal behavior to achieve interaction goals, namely body motion ~\cite{2011sanghvi}, facial expressions ~\cite{2013cid}, or human gaze ~\cite{kontogiorgos2021systematic}. Challenges to the more widespread use of human social cues in HRI include the subtlety of these signals and the absence of reactions to errors in certain contexts ~\cite{2015mirnig}. Other challenges include the incongruence that may emerge in human nonverbal reactions to robot failure ~\cite{2016hayes}. ~\citet{giuliani2015systematic} found effects in the nature of the failure and the surroundings on human nonverbal responses to robot failure. First, people tend to talk more if the robot failure consists of a social norm violation than if it is a technical failure. Second, people exhibit fewer nonverbal signals, such as smiling, nodding, and head shaking, if no other people are present. ~\citet{2019morales} also found that human reactions to robot failures vary with robot appearance. Anthropomorphism of robots has also been seen to impact reactions ~\cite{2020kontogiorgosembodiment}. These variations in human responses to robot failure call for carefully developed methods that can be generalized.

People are responsive to robots' social cues ~\cite{breazeal2005effects}; robots, in turn, should also be equipped with strategies that allow them to leverage users' social cues to achieve better task performance. Interesting work has used these types of inputs in reinforcement learning systems ~\cite{hwang2014reward}. Future work should further explore the potential of human nonverbal inputs to achieve continuous robot learning and improve task completion.

\section{Theme 3: Machine learning methods for failure detection in HRI with social cues}
\label{sec:theme3}

This next section on state-of-the-art algorithmic techniques is mostly relevant to those developing practical applications. We aim to provide an overview of technical approaches as a starting point to help researchers navigate the space of potential solutions -- as such, this section will have a slightly different structure than the prior two themes. No recent review has investigated the technical approaches toward social cue recognition to resolve HRI task failures. Here, we bring together the existing Machine Learning (ML) approaches related to detecting task failures, including intention detection, human action classification, and social cue recognition (see \autoref{tab:approaches}). We will dive into prior work surrounding the following questions:

\begin{itemize}
\item What are the technical approaches to failure detection in HRI?
\item What are some potentially useful machine learning approaches?
\item What is needed to employ these technical approaches?
\end{itemize}
 
\subsection{How is failure detection typically approached from a technical standpoint?}

\subsubsection{Failure detection}
Failure detection in human-human and human-robot interactions has historically been approached using rule-based and heuristic models \cite{askarpour2017modeling,askarpour2019formal}, such as task analysis. In task analysis, errors are manually annotated \cite{baber2002task}. New ML tools provide automated classification and identification opportunities. \citet{kulsoom2022review} reviewed human activity detection and recognition approaches, including machine learning, reinforcement learning, transfer learning, and deep learning -- identifying data collection and generalizability challenges. However, support vector machines seem robust to decreases in performance accuracy as long as the dataset is large. Deep learning classifiers are favored for learning quickly on complex datasets, like in healthcare use cases. \citet{kong2022human}'s review on state-of-the-art technical approaches in action recognition and prediction points out several challenges, namely variations between and within data classes, uncontrolled outdoor environments, lack of annotated datasets, hierarchies of complex actions, and uneven predictability across video frames. These issues are also present in HRI -- often, humans and robots collaborate in uncontrolled environments, datasets might be scarce, and the actions between humans and robots may be incredibly complex.

\subsubsection{Interpreting reactions}
To approach the modeling and interpretation of complex human reactions to failure, it can be useful to consider methods in affective computing. Here, Machine learning (ML) and Deep Learning (DL) methods have long been used to predict emotional states \cite{wang2022systematic}. Emotional cues are important for human-human interaction, and representations of these emotional states can help robots respond better to failures. Facial expressions have been collected, codified, and used extensively in emotion research \cite{ekman1978facial,ekman1992there}, and in HRI failures \cite{kontogiorgos2021systematic}. Datasets of human expressions are available, including \citet{sun2020eev}'s Evoked Expressions from Videos (EEV) dataset. \citet{zhang2014bp4d} released a dataset of spontaneous (i.e., not posed) facial expressions in \textit{3D videos} -- filling a gap in the affective computing literature, which is dominated by posed, 2D imagery. They include facial actions, head/pose data, and landmarks in 2D and 3D. Other work contributes datasets of human activity \cite{kay2017kinetics, epstein2020oops}. The Oops! dataset \cite{epstein2020oops}, is used to identify intentional and unintentional actions. Beyond facial expressions, body motion can also be used for identifying intention detection, including gestures \cite{xiao2014human} and body poses \cite{mccoll2011human}. 

\begin{table}[ht]
\resizebox{\textwidth}{!}{%
\begin{tabular}{lllll}
\hline
Reference &
  Task &
  Datasets &
  Modalities of Data &
  Modeling Approach \\ \hline
Das et al., 2021 \cite{das2021plsm} &
  \begin{tabular}[c]{@{}l@{}}Human activity recognition\\ (intentional vs. unintentional)\end{tabular} &
  Oops! \cite{epstein2020oops} &
  In-the-wild videos &
  Parallelized Liquid State Machine \\ \hline
Epstein et al., 2020 \cite{epstein2020oops} &
  \begin{tabular}[c]{@{}l@{}}Human activity recognition\\ (intentional vs. unintentional)\end{tabular} &
  Oops! &
  In-the-wild videos &
  \begin{tabular}[c]{@{}l@{}}3D-CNN*\\ Kinetics \cite{kay2017kinetics}\end{tabular} \\ \hline
\begin{tabular}[c]{@{}l@{}}Epstein \& Vondrick, \\ 2021 \cite{epstein2021learning}\end{tabular} &
  \begin{tabular}[c]{@{}l@{}}Human activity localization \\ (intentional vs. unintentional)\end{tabular} &
  Oops! \cite{epstein2020oops} &
  Annotated in-the-wild videos &
  \begin{tabular}[c]{@{}l@{}}3D CNN + Attention-based transformers*\\ Kinetics finetuned \cite{kay2017kinetics}\\ Kinetics\\ 3D CNN only\end{tabular} \\ \hline
Jenni et al., 2020 \cite{jenni2020video} &
  Human activity recognition &
  \begin{tabular}[c]{@{}l@{}}Kinetics \cite{kay2017kinetics}\\ UCF101 \cite{soomro2012ucf101}\\ HMDB51 \cite{kuehne2011hmdb}\end{tabular} &
  In-the-wild videos &
  3D-ResNet18, C3D \\ \hline
\begin{tabular}[c]{@{}l@{}}Nwakanma \\ et al., 2021 \cite{nwakanma2021detection}\end{tabular} &
  \begin{tabular}[c]{@{}l@{}}Human activity recognition\\ (normal vs. abnormal)\end{tabular} &
  Collected their own &
  \begin{tabular}[c]{@{}l@{}}Vibration, Respiration, \\ Movement, LIDAR\end{tabular} &
  \begin{tabular}[c]{@{}l@{}}CNN*\\ kNN\\ Naive Bayes\\ Logistic Regression\\ SVM\end{tabular} \\ \hline
\begin{tabular}[c]{@{}l@{}}Ramos de Assis Neto \\ et al., 2020 \cite{de2020detecting}\end{tabular} &
  Human activity recognition &
  UP-Fall Detection \cite{martinez2019up} &
  \begin{tabular}[c]{@{}l@{}}(3-axis accelerometer \& \\gyroscope, EEG, \\ambient light), \\ infrared sensors, video\end{tabular} &
  Bi-LSTM \\ \hline
\begin{tabular}[c]{@{}l@{}}Synakowski \\ et al., 2021 \cite{synakowski2021adding}\end{tabular} &
  \begin{tabular}[c]{@{}l@{}}Human activity recognition\\ (intentional vs. unintentional)\end{tabular} &
  Collected their own &
  \begin{tabular}[c]{@{}l@{}}3D Trajectories (center of mass)\\ 3D Human joints location\\ 3D Human pose\end{tabular} &
  Unsupervised computer vision \\ \hline
Xu et al., 2022 \cite{xu2022probabilistic} &
  \begin{tabular}[c]{@{}l@{}}Human activity localization\\ (intentional vs. unintentional)\end{tabular} &
  Oops! \cite{epstein2020oops} &
  In-the-wild videos &
  Dense Probabilistic Localization \\ \hline
Zhou et al., 2021 \cite{zhou2021temporal} &
  \begin{tabular}[c]{@{}l@{}}Human activity localization\\ (intentional vs. unintentional)\end{tabular} &
  Oops! \cite{epstein2020oops} &
  In-the-wild videos &
  Temporal Probabilistic Regression \\ \hline
\begin{tabular}[c]{@{}l@{}}Vinanzi et al., 2021 \\ \cite{vinanzi2021collaborative}\end{tabular} &
  Human intention detection &
  Collected their own &
  Body pose, gaze &
  Feature-Space Split Clustering \\ \hline
\begin{tabular}[c]{@{}l@{}}Addo \& Ahamed, \\ 2014 \cite{addo2014applying}\end{tabular} &
  Affect recognition &
  Collected their own &
  \begin{tabular}[c]{@{}l@{}}Facial expressions, \\ speech\end{tabular} &
  Reinforcement learning (greedy) \\ \hline
Shi et al., 2019 \cite{shi2019automatic} &
  Academic confusion &
  Collected their own &
  Facial expressions &
  \begin{tabular}[c]{@{}l@{}}HOG-SVM\\ LBP-SVM\\ CNN\\ CNN-SVM*\end{tabular} \\ \hline
\begin{tabular}[c]{@{}l@{}}Srinivasa et al., \\ 2017 \cite{srinivasa2017analysis}\end{tabular} &
  Facial expressiveness &
  Affectiva-MIT \cite{mcduff2013affectiva} &
  Facial expressions &
  LSTM \\ \hline
Sun et al., 2020 \cite{sun2020eev} &
  Evoked expressions &
  EEV \cite{sun2020eev} &
  Facial expressions &
  LSTM \\ \hline
Zhang et al., 2021 \cite{zhang2021real} &
  Emotion recognition &
  \begin{tabular}[c]{@{}l@{}}IEMOCAP \cite{busso2008iemocap}\\ MELD \cite{poria2018meld}\end{tabular} &
  \begin{tabular}[c]{@{}l@{}}Text, video \& audio \\ (dyadic conversation)\end{tabular} &
  \begin{tabular}[c]{@{}l@{}}ERLDK*\\ context-LSTM + Att\\ Memory Fusion Network\\ Tensor Fusion Network\\ Conversational Memory Network\\ Interactive Conversational Memory Network\\ BiDialogueRNN + Att\end{tabular} \\ \hline
\begin{tabular}[c]{@{}l@{}}Ben-Youssef et al., 2021\\ \cite{ben2019early}\end{tabular} &
  \begin{tabular}[c]{@{}l@{}}Engagement breakdown\\ (engagement vs. disengagement)\end{tabular} &
  UE-HRI \cite{Ben-Youssef:2017:UE-HRI_Dataset} &
  \begin{tabular}[c]{@{}l@{}}Sonar, Laser, Gaze,\\ Head position,\\ Facial expression,\\ speech\end{tabular} &
  Logistic Regression \\ \hline
\begin{tabular}[c]{@{}l@{}}Kontogiorgos \\ et al., 2020 \cite{kontogiorgos2020behavioural}\end{tabular} &
  \begin{tabular}[c]{@{}l@{}}Failure recognition \\ (HRI cooking task)\end{tabular} &
  Collected their own &
  \begin{tabular}[c]{@{}l@{}}Gaze, head movement, \\ speech\end{tabular} &
  Random Forest Classifier \\ \hline
Stiber et al., 2022 \cite{2022stiber} &
  \begin{tabular}[c]{@{}l@{}}Robot error detection\\ (error vs. no error)\end{tabular} &
  Collected their own &
  Facial expressions &
  \begin{tabular}[c]{@{}l@{}}Weighted binary classification +\\ Sliding window filtering\end{tabular} \\ \hline
Trung et al., 2017 \cite{2017trung} &
  \begin{tabular}[c]{@{}l@{}}Robot error detection \\ (social norm violations vs.\\ technical failures)\end{tabular} &
  Collected their own &
  \begin{tabular}[c]{@{}l@{}}Head and shoulder movement,\\ body movement\end{tabular} &
  \begin{tabular}[c]{@{}l@{}}Rule learner\\ kNN\\ Naive Bayes\end{tabular} \\ \hline
Li et al., 2020 \cite{li2020facial} &
  Infinite Mario performance &
  Collected their own &
  \begin{tabular}[c]{@{}l@{}}Facial expressions, \\ positive \& negative feedback\end{tabular} &
  TAMER (Reinforcement Learning) \cite{knox2009interactively} \\ \hline
\end{tabular}%
}
\caption{State-of-the-art algorithmic approaches for social cue recognition.\protect\footnotemark Approaches marked with an asterisk (*) are bench-marked against other listed approaches.}

\label{tab:approaches}
\end{table}
\newpage
\footnotetext{CNN: Convolutional Neural Network; C3D: 3D Convolutional Network; LIDAR: Light Detection and Ranging; kNN: k-Nearest Neighbors; SVM: Support Vector Machine; EEG: Electroencephalography; LSTM: Long Short-Term Memory; HOG-SVM: Histogram of Oriented Gradient + SVM; LBP-SVM: Local Binary Patterns + SVM; ERLDK: Emotion Reinforcement Learning and Domain Knowledge; Att: Attention; TAMER: Training an Agent Manually via Evaluative Reinforcement.}

\subsection{What are some promising machine learning approaches?}
\subsubsection{Unsupervised learning approaches: when data isn't labeled}
Unsupervised learning is one approach for recognizing implicit human social cues \cite{martinez2016advances, carbonell2021comparing}. Common unsupervised approaches include Self-Supervised Learning (SSL) methods, which generate labels from unstructured data, assign labels to the data, and use the self-generated labels to continue training themselves. The ground-truth labels change with each training iteration. Other approaches involve the use of probability distributions through the use of neural networks and probabilistic regression. 

\citet{epstein2020oops} first used a self-supervised method, relying on withholding some of the video input, using natural data features. Compared to a baseline model trained on the Kinetics action recognition dataset \cite{2017carreirakinetics}, \citet{epstein2020oops}'s 3D Convolutional Neural Network (CNN) model performed similarly while only using video speed as a predictive feature. \citet{epstein2021learning} later added video annotations of short descriptions of the goals and failures of each video with a decoder, resulting in better performance than a Kinetics-trained model. \citet{zhou2021temporal} applied Long short-term memory (LSTM) modeling -- a type of Recurrent Neural Network (RNN) that captures the temporal dependencies of continuous signals.
Instead of training LSTM model on hard labels, \citet{zhou2021temporal} created probability distributions from video annotations and aggregated the distributions with an online label attention model. 

The resulting online model was more accurate at localizing intentionality than offline methods which relied on gross movement features. \citet{xu2022probabilistic} took an alternative approach, using Dense Probabilistic Localization and temporal label aggregation for unintentional action localization. Similar to \citet{zhou2021temporal}, they generated labels with probabilistic annotation modeling and then trained their model using three different dense supervision techniques: probabilistic dense classification, probabilistic temporal detection, and probabilistic regression. 

Other approaches include \citet{synakowski2021adding} who calculated 3D kinematics, self-propelled motion, and Newtonian motion and their combined relationship to determine intentionality. \citet{synakowski2021adding} present three datasets: \textit{intent-maya} (3D animations of objects), \textit{intent-mocap} (motion capture videos of humans without center-of-mass information), and \textit{intent-youtube} (in-the-wild videos of humans performing actions). Using motion and extracted kinematics features as inputs, their model outperformed other ML models when classifying whether actions were intentional or unintentional. Extracted kinematic information may thus be a promising additional feature for unsupervised intentionality prediction. A Hidden Markov Model model proposed by \citet{scheirer2002frustrating} combined human physiological data (galvanic skin response (GSR) and blood pressure) and behavioral data (mouse clicks) to detect frustration when participants played a game that purposefully introduced delays to cause frustration. 

\citet{vinanzi2021collaborative} combined gaze and body pose in a Feature-Space Split Clustering model. A robot used this model to predict whether a human's intentions would lead to task success in a collaborative block-building task. 

In overall, unsupervised learning algorithms can be most suitable for failure detection when social cue data is unlabeled. These methods have been used to detect intent and predict task success. New opportunities in HRI research include obtaining social cues from unstructured "in-the-wild" data, using new analysis methods to extract insights from these data, and automating labeling through self-supervised learning methods.

\subsubsection{Supervised learning approaches: when data is labeled}

Supervised learning algorithms have been used for action recognition, such as recognition of human postures \cite{ghazal2019human, wang2016human}, and daily activities \cite{sagha2011benchmarking}. Unlike unsupervised approaches, supervised methods rely on labeled data. While labeling data is often expensive, labeled datasets for human activity detection exist, including fall detection (UP-Fall Detection \cite{martinez2019up}), human activity datasets (Kinetics with 400 classes \cite{kay2017kinetics}; UCF101 with 101 classes \cite{soomro2012ucf101}; HMDB51 with 51 classes \cite{kuehne2011hmdb}), and facial expression datasets (Affectiva-MIT \cite{mcduff2013affectiva}; see \citet{li2020deep} for a survey on facial expression datasets). Common implementations of supervised ML in human activity recognition include support vector machines (SVM), k-Nearest Neighbors (kNN), and random forest classifiers. 

Various data modalities have been considered for supervised learning methods. \citet{attal2015physical} review techniques with human wearable sensor data. Some of these works include \citet{de2020detecting} who detected falls and six other daily activities with a bidirectional LSTM -- processing both past and future information at each time step. Nwakanma and colleagues \cite{nwakanma2021detection} tested CNNs against other supervised ML approaches to predict the efficacy of an emergency detection system within a smart factory. Similar to \citet{de2020detecting}, data was multi-modal and included light detection and ranging (LIDAR), breathing patterns, and vibration patterns. The CNN outperformed all other models, reaching a classification accuracy of at least 99\% on per modality. However, the fusion of multimodal data at different time points was challenging. 

Facial expressions are commonly used for social cue recognition, with facial action units (AUs) being well-studied across psychology. \citet{shi2019automatic} compared approaches for facial expression recognition, including supervised ML approaches like HOG-SVM (Histogram of Oriented Gradient Support Vector Machine) and LBP-SVM (Local Binary Patterns Support Vector Machine), a deep learning approach with CNNs, and their combined approach using CNN-SVM. The authors aimed to predict confused states from facial expressions. The combined CNN-SVM had the best predictive performance. \citet{zeng2004bimodal} developed a bimodal fusion method for affect recognition, combining AUs and speech prosody information, outperforming a unimodal method. Recently, \citet{ben2019early} conducted an HRI study fusing multimodal data (gaze, distance, speech, facial expressions, head position) to predict task failures (i.e., disengagement from a task) using logistic regression. A combination of distance from the robot, facial expressions, head position, gaze, and speech produced the best predictions for task failure up to 10 seconds before it occurs. \citet{2018short} evaluated contingency (i.e., how the environment (human interactant) reacts to a robot's action)  using audio and visual data features. 

Some works are particularly relevant to leveraging human social cues to detect failure. \citet{kontogiorgos2020behavioural} used multimodal data (gaze, head movement, speech, and reaction times) in a Random Forest classifier to help a robot detect conversational failure. \citet{2021shi} describes a method to detect user intention through the user's gaze. \citet{kontogiorgos2021systematic} used an instruction corpus (participants guided by a robot in a cooking task) and a negotiation corpus (participants negotiating with a robot in a decision-making task) to implement a failure prediction model. Lexical features (tone, affect, positive and negative emotion) were highly significant in predicting task failure, and multi-modal information streams performed better for failure classification. \citet{2022stiber} used facial expressions to detect and localize robot errors in HRI. The authors built a two-stage model: 1) weighted binary classification and 2) filtering through a sliding window. Interestingly, the authors observed changing facial expressions in \textit{anticipation} of certain robot errors. 

In short, a significant challenge exists in multimodal fusion between different sensors, timescales, data distributions, and complexity of use cases. However, despite the technical challenge of fusing multimodal data, supervised machine learning models that incorporate multimodal information streams as predictors of activity, interaction, or failure often outperform models with a unimodal predictor. 

\subsubsection{Reinforcement learning approaches: iterative learning}

Reinforcement Learning (RL) is a less explored space regarding social cue-based failure detection in HRI. In RL, an agent makes decisions based on the effect of an action on the environment, learning iteratively \cite{sutton1998reinforcement, sutton2018reinforcement}. We will here provide an introduction that can be used to better contextualize the research space of RL for this use. 

\citet{akalin2021reinforcement} reviewed reinforcement learning methods in social robotics, discovering three themes: \textit{interactive reinforcement learning}, where humans provide feedback during the learning process; \textit{intrinsically motivated methods}, where the robot considers internal and external dynamics; and \textit{task performance driven methods}, where the reward depends on human performance, robot performance, or both. \textit{Interactive reinforcement learning} requires human feedback that is understandable to a robot. However, advantages include a more personalized and natural adaptation of the agent's behavior.

Different data types can inputs for reinforcement learning-based systems. \citet{lin2020review} reviewed human feedback for interactive robotic systems. Common forms of feedback include mouse clicks or keystrokes, which aren't how people naturally provide feedback. \citet{addo2014applying} tested multimodal feedback in HRI with social robot 'ZOEI' who entertained a crowd through stand-up comedy. Participants would respond naturally and explicitly state how funny ZOEI's joke was. Qualitative findings indicated that ZOEI managed to tell funnier jokes as it learned from social cues. \citet{weber2018shape} performed a similar study where their Reeti robot used facial expressions (smiles and grimaces) and speech patterns (laughs) and reinforcement learning (Q-learning) to improve engagement and humor. \citet{li2020facial} implemented the model TAMER (Training an Agent Manually via Evaluative Reinforcement) using facial expressions and feedback from human operators. TAMER learns not from a pre-programmed reward function, but from real-time human interactions, achieving success in various simulated tasks \cite{knox2009interactively,knox2012reinforcement}. \citet{li2020facial} found that incorporating overt facial expressions into a RL model enables the agent to succeed in the game of Infinite Mario. \citet{knox2013training} implemented the TAMER RL model into a physical, social robot (Nexi), which was able to learn five different behaviors (go-to, keep conversational distance, look away, toy tantrum, and magnetic control). It took from 4.7 -- 27.3 minutes of active training time per behavior. By having robots learn from expressed human behaviors, whether explicit or implicit, reinforcement learning is a promising option for successful HRI, especially in cases where initial errors are more forgivable, so that a robot has time to learn. 

In short, interactive reinforcement learning in HRI requires ensuring that human feedback can be interpreted by the robot. Reinforcement learning algorithms can enable feedback and interaction between humans and robots that are more natural than those based on heuristic systems.

\subsection{What is needed to apply these technical approaches?}
Applications of ML methods in human emotion, affect, and activity recognition might be extrapolated to the context failure detection in HRI. Common inputs include audio and visual modalities -- facial expressions being especially widely used. However, multimodal streams of information, such as human wearable sensor data, can outperform unimodal models in certain contexts, despite challenges with multimodal models. Many works use facial expressions for recognizing phenomena such as confusion \cite{shi2019automatic}, affect \cite{addo2014applying}, and expressiveness \cite{srinivasa2017analysis}. \citet{ben2019early}'s work is closer to our proposed work by using a multimodal human data stream to predict disengagement in a task, which is here analogous to task \textit{failure}.

Much prior work proposes models by comparing their performance on a dataset (such as Oops! \cite{epstein2020oops}) against other algorithmic approaches \cite{shi2019automatic,nwakanma2021detection,zeng2004bimodal,zhang2021real,epstein2021learning}, rather than against an online, real-time deployment of these models. It is difficult to systematically assess the most optimal approach to a research goal (e.g., detecting task failure through social cue recognition), given that models are often context-specific and dependent on the type of training data. 

The applicability of each technical approach thus depends on the use case. Are robots equipped with sensors to perceive their surroundings? Or are the sensors external, such as external cameras or data from wearable sensors worn by people? Existing work shows that facial responses can used be for error detection in robots \cite{2022stiber}. However, many current works use laboratory-controlled datasets. Collecting data in naturalistic scenarios (see, for example, \citet{bremers2023bystander}) is a necessary step for determining the feasibility of these systems in-the-wild. By providing an overview of the relevant existing work in this and related spaces, we aim to provide a starting point for research teams addressing the technical challenges of failure detection based on social signals for HRI. 

\subsection{In summary}
\label{subsubsec:datasets}
The Oops! dataset ~\cite{epstein2020oops} provides a starting point for modeling reactions to failures. A research opportunity lies in expanding Oops! to include videos of \textit{robots} making mistakes. Many databases of facial expressions exist for experimentation on new facial recognition and affective response models (RAF-DB ~\cite{li2017reliable}, CK+ ~\cite{lucey2010extended}, Oulu-CASIA ~\cite{zhao2011facial}, AFEW 7.0 ~\cite{dhall2012collecting}, BP4D-Spontaneous ~\cite{zhang2014bp4d}); however, sparse datasets exist for facial expressions across HRI contexts (such as \cite{bremers2023bystander}). Further, some research indicates that variances exist between male and female facial expressions ~\cite{zeng2004bimodal} and cultural differences in facial reactions to their environment ~\cite{jack2012facial} although this is debated~\cite{hwang2015evidence}. More research is needed on social cue recognition models that can account for demographic variables (for a systematic approach using deep learning techniques, see ~\citet{fan2020facial}). Finally, we mention ~\citet{2020aneja}'s Agent Conversational Error (ACE) dataset with transcripts and error annotations. 

Practitioners in this space should ask themselves the following questions when choosing a technical approach:

\begin{itemize}
\item What is the type of failure?
\item What are available data inputs? (including natural and extracted data streams)
\item What resources will you have for labeling data? (e.g., computational resources, human assistance)
\item What is the ideal trade-off between precision and recall? (i.e., in some settings, failure detection is critical)
\item Does detection need to happen in real-time?
\end{itemize}

Researchers can choose to start with models that have been tested in similar conditions to their context. For example, for unintentional action localization or detection, many self-supervised and unsupervised learning approaches are used in conjunction with the Oops! dataset ~\cite{epstein2020oops}. Failure detection that leans on affect and emotion recognition may rely more heavily on supervised learning methods, which have well-established indicators through defined facial action units that make labeling efficient. 

Multi-modal data input approaches often outperform uni-modal approaches, especially in activity recognition, affect recognition, and engagement ~\cite{zeng2004bimodal,ben2019early,de2020detecting}. However, this should be systematically investigated across multiple use cases, as the literature on technical approaches for failure detection in HRI with social cues is sparse.

Large, multi-modal models can be expensive to train, especially for implementations at scale. Cost-aware pre-training approaches for deep learning architectures can make these algorithms more accessible ~\cite{chung2015cost} -- these have been used in HRI ~\cite{2021liau}. Another common bottleneck is manual data labeling, which is time-intensive and requires multiple annotators. There are two potential alternatives: the use of labeled datasets, as discussed in Section \ref{subsubsec:datasets}, or the use of self-supervised learning and probabilistic labeling algorithms, which may increase noise in data labels.

In some situations, detection of failures in HRI is time-intensive and safety-critical. A survey of ML algorithms in industrial settings ~\cite{mukherjee2022survey} reports that accuracy in the deployed models needs to be high to allow for integration in the working environment. This, however, could lead to issues with model overfitting. Thus, the deployment of models in critical settings requires a performance verification and testing process. In other scenarios where model errors can be permitted, reinforcement learning algorithms may enable robots to learn from social cues in real-time. The ML field is advancing rapidly, and we anticipate the implementation of these algorithmic approaches for HRI in working and collaborative environments will not be long.

\section{Discussion}
\label{sec:discussion}

\subsection{General discussion} 
The current state-of-the-art reveals promises to achieve socially self-aware robots for failure detection, yet gaps remain to be addressed in future research. Based on our discussion of the three central research themes, we make the following recommendations to HRI researchers and practitioners interested in using social cues to identify robot task failures.

First, to overcome the lack of a single unified definition of robot error, researchers can draw from the available definitions and taxonomies of human and robot errors to help them describe rather than define errors at hand. We suggest adopting the definition of error as "actions not as intended" \cite{hollnagel1991phenotype} -- refraining from further definition and acknowledging the richness and complexity each instance of an error entails. 

Second, seamless deployment of ML-based models may be better accomplished in data-rich environments that are already monitored, like industrial settings ~\cite{nwakanma2021detection} and healthcare settings ~\cite{johanson2021improving, cao2017survey}. However, other factors could influence the feasibility and desirability of applications in these fields. For instance, in the medical field, privacy, legal protections, and participants' willingness should be the core considerations. 

Third, from a technical perspective, video datasets of human facial reactions can be helpful starting points even if subjects are not reacting to robot failures or even failures at all -- as long as representativeness is kept in mind. Unsupervised and self-supervised algorithms may be particularly beneficial when there is a lack of labeled data, as is often the case for data collection in the wild. In settings where failures during deployment are more acceptable, reinforcement learning algorithms are a promising approach where robots can learn from direct interactions. Multi-modal approaches tend to outperform unimodal approaches. However, practitioners should be aware of issues related to integrating multiple data types. 

Fourth, regarding reporting standards, we recommend that ML practitioners in social robotics benchmark new algorithmic approaches against state-of-the-art models and report general accuracy metrics along with other specifically applicable performance metrics. Performance across different data folds or seeds should also be reported to evaluate robustness. Other valuable metrics that can be included in reports are Cohen's Kappa ~\cite{cohenskappa}, to account for data labeling noise (agreeability between labels predicted by the model and original labels), as well as balanced accuracy ~\cite{accbalance}, to account for data imbalance when calculating performance.

\subsection{Future work: a research agenda}

\subsubsection{Dataset collection and availability}
Increasing the availability of datasets could be very beneficial in advancing this research direction. ~\citet{2015mirnig} collected a video dataset of humans interacting with robots across multiple studies, but this was not publicly available. ~\citet{kontogiorgos2021systematic} also collected a rich corpus of data that was not released. Just like ~\citet{epstein2020oops}'s Oops! dataset on human unintentional actions led to significant advances in the field of action and intentionality detection ~\cite{epstein2021learning,das2021plsm,zhou2021temporal, xu2022probabilistic}, we believe that the public release of such datasets could help accelerate the development of this research direction through shared resources and collaboration between interdisciplinary research groups. As much of these datasets will involve identifiable information on human subjects (namely, their face), we recommend that clarifying privacy and data usage guidelines can mitigate some risk of public dataset release and ensure datasets are used for their intended purposes, as opposed to other types of (undesired) ML use cases. We advise researchers to take caution when releasing data and incorporate measures such as the requirement of Institutional Review Boards (IRBs) or ethical approval, an affiliated institution, and the provision of a protocol before releasing large datasets of human subjects data to interested third parties. This could, in part, be achieved through existing measures of data repositories, such as the Qualitative Data Repository \cite{qdr}, which have prior experience managing sensitive data. 

\subsubsection{Considering types of social cues and types of interactants}
Multi-modal approaches tend to perform better than unimodal approaches. New research could investigate adding data streams that are less frequently considered, such as body pose, speech prosody and bio-signals, and signals from wearable sensors. 

In this review, we took a broad approach toward social signals that encompasses both dyadic interactants and bystanders. Future work could address specific interaction scenarios that occur in operator-robot interactions. For instance, some research has looked into the particular feedback that operators and dyadic interactants can provide, which adds the consideration as to whether the person's reaction is a reaction to the robot or a reaction to their \textit{own} action \cite{candon2023nonverbalhuman,candon2023nonverbal,candon2024leveraging}. On a related note, temporal changes can also be considered. For instance, \citet{parreira2023badidea} propose using anticipatory reactions to predict whether a robot might be committing a failure soon.

\subsubsection{From the lab to the real world}
Bringing findings from HRI from the lab into the real world brings along its own challenges that include both higher complexity of interactions and an increased technical complexity \cite{bu2024field}. However, to achieve domain adaptability, it should be kept in mind when models are trained on lab data and released in the wild \cite{parreira2024study}. In the case of socially-aware robots, there is a need to consider the implications of data analysis and storage when it comes to human signals that could be identifiable.

It will also be essential to test the effect of demographic features, such as age, gender, personality, (dis)ability, or cultural background, on how humans recognize and react to robot failures, as some work indicates that robot acceptance, collaboration, and trust are affected ~\cite{2014salem,2015salem}. Other works point out expectations brought about by artificial agents' anthropomorphic appearance and embodiment ~\cite{2001bickmore,Deng_2019}, and the effect these may have in human responses to failure ~\cite{kontogiorgos2020behavioural}. This calls for further exploration. 

\section{Conclusion}
There is a clear need for robots to be able to adapt to complex and dynamic environments. We outline the human ability to detect errors based on observation of social cues and highlight how human-robot interaction could benefit from applying this concept to robotics. We highlight the current state-of-the-art on this topic and review applicable technical methods to achieve social cue-based error recognition. Finally, we propose a research agenda based on gaps in the literature. Failure detection for HRI through human social cues remains a field with much potential to be explored, with multiple application opportunities and growing pertinence. 

\section{Acknowledgements}
We thank members of the Cornell Tech and Accenture Labs research communities for their feedback on earlier versions of this work -- in particular: Michael Kuniavsky, Manaswi Saha, Adolfo Ramirez-Aristizabal, Mirjana Spasojevic, and Natalie Friedman. This research was conducted as part of a collaboration funded by Accenture Labs.

\bibliographystyle{ACM-Reference-Format}
\bibliography{bibliography.bib}


\begin{thebibliography}{166}


\ifx \showCODEN    \undefined \def \showCODEN     #1{\unskip}     \fi
\ifx \showDOI      \undefined \def \showDOI       #1{#1}\fi
\ifx \showISBNx    \undefined \def \showISBNx     #1{\unskip}     \fi
\ifx \showISBNxiii \undefined \def \showISBNxiii  #1{\unskip}     \fi
\ifx \showISSN     \undefined \def \showISSN      #1{\unskip}     \fi
\ifx \showLCCN     \undefined \def \showLCCN      #1{\unskip}     \fi
\ifx \shownote     \undefined \def \shownote      #1{#1}          \fi
\ifx \showarticletitle \undefined \def \showarticletitle #1{#1}   \fi
\ifx \showURL      \undefined \def \showURL       {\relax}        \fi
\providecommand\bibfield[2]{#2}
\providecommand\bibinfo[2]{#2}
\providecommand\natexlab[1]{#1}
\providecommand\showeprint[2][]{arXiv:#2}

\bibitem[Addo and Ahamed(2014)]%
        {addo2014applying}
\bibfield{author}{\bibinfo{person}{Ivor~D Addo} {and} \bibinfo{person}{Sheikh~I Ahamed}.} \bibinfo{year}{2014}\natexlab{}.
\newblock \showarticletitle{Applying affective feedback to reinforcement learning in ZOEI, a comic humanoid robot}. In \bibinfo{booktitle}{\emph{The 23rd IEEE International Symposium on Robot and Human Interactive Communication}}. IEEE, \bibinfo{publisher}{IEEE}, \bibinfo{address}{New York, NY, USA}, \bibinfo{pages}{423--428}.
\newblock


\bibitem[Akalin and Loutfi(2021)]%
        {akalin2021reinforcement}
\bibfield{author}{\bibinfo{person}{Neziha Akalin} {and} \bibinfo{person}{Amy Loutfi}.} \bibinfo{year}{2021}\natexlab{}.
\newblock \showarticletitle{Reinforcement learning approaches in social robotics}.
\newblock \bibinfo{journal}{\emph{Sensors}} \bibinfo{volume}{21}, \bibinfo{number}{4} (\bibinfo{year}{2021}), \bibinfo{pages}{1292}.
\newblock


\bibitem[Aneja et~al\mbox{.}(2020)]%
        {2020aneja}
\bibfield{author}{\bibinfo{person}{Deepali Aneja}, \bibinfo{person}{Daniel McDuff}, {and} \bibinfo{person}{Mary Czerwinski}.} \bibinfo{year}{2020}\natexlab{}.
\newblock \showarticletitle{Conversational Error Analysis in Human-Agent Interaction}. In \bibinfo{booktitle}{\emph{Proceedings of the 20th ACM International Conference on Intelligent Virtual Agents}} (Virtual Event, Scotland, UK) \emph{(\bibinfo{series}{IVA '20})}. \bibinfo{publisher}{Association for Computing Machinery}, \bibinfo{address}{New York, NY, USA}, Article \bibinfo{articleno}{3}, \bibinfo{numpages}{8}~pages.
\newblock
\showISBNx{9781450375863}
\urldef\tempurl%
\url{https://doi.org/10.1145/3383652.3423901}
\showDOI{\tempurl}


\bibitem[Aronson(2018)]%
        {2010aronson}
\bibfield{author}{\bibinfo{person}{Reuben~M. Aronson}.} \bibinfo{year}{2018}\natexlab{}.
\newblock \showarticletitle{Gaze for Error Detection During Human-Robot Shared Manipulation}. In \bibinfo{booktitle}{\emph{RSS Workshop: Towards a Framework for Joint Action}}.
\newblock


\bibitem[Askarpour et~al\mbox{.}(2017)]%
        {askarpour2017modeling}
\bibfield{author}{\bibinfo{person}{Mehrnoosh Askarpour}, \bibinfo{person}{Dino Mandrioli}, \bibinfo{person}{Matteo Rossi}, {and} \bibinfo{person}{Federico Vicentini}.} \bibinfo{year}{2017}\natexlab{}.
\newblock \showarticletitle{Modeling operator behavior in the safety analysis of collaborative robotic applications}. In \bibinfo{booktitle}{\emph{International Conference on Computer Safety, Reliability, and Security}}. Springer, \bibinfo{publisher}{Springer}, \bibinfo{address}{Cham}, \bibinfo{pages}{89--104}.
\newblock


\bibitem[Askarpour et~al\mbox{.}(2019)]%
        {askarpour2019formal}
\bibfield{author}{\bibinfo{person}{Mehrnoosh Askarpour}, \bibinfo{person}{Dino Mandrioli}, \bibinfo{person}{Matteo Rossi}, {and} \bibinfo{person}{Federico Vicentini}.} \bibinfo{year}{2019}\natexlab{}.
\newblock \showarticletitle{Formal model of human erroneous behavior for safety analysis in collaborative robotics}.
\newblock \bibinfo{journal}{\emph{Robotics and computer-integrated Manufacturing}}  \bibinfo{volume}{57} (\bibinfo{year}{2019}), \bibinfo{pages}{465--476}.
\newblock


\bibitem[Attal et~al\mbox{.}(2015)]%
        {attal2015physical}
\bibfield{author}{\bibinfo{person}{Ferhat Attal}, \bibinfo{person}{Samer Mohammed}, \bibinfo{person}{Mariam Dedabrishvili}, \bibinfo{person}{Faicel Chamroukhi}, \bibinfo{person}{Latifa Oukhellou}, {and} \bibinfo{person}{Yacine Amirat}.} \bibinfo{year}{2015}\natexlab{}.
\newblock \showarticletitle{Physical human activity recognition using wearable sensors}.
\newblock \bibinfo{journal}{\emph{Sensors}} \bibinfo{volume}{15}, \bibinfo{number}{12} (\bibinfo{year}{2015}), \bibinfo{pages}{31314--31338}.
\newblock


\bibitem[Baber and Stanton(2002)]%
        {baber2002task}
\bibfield{author}{\bibinfo{person}{Chris Baber} {and} \bibinfo{person}{Neville~A Stanton}.} \bibinfo{year}{2002}\natexlab{}.
\newblock \showarticletitle{Task analysis for error identification: theory, method and validation}.
\newblock \bibinfo{journal}{\emph{Theoretical Issues in Ergonomics Science}} \bibinfo{volume}{3}, \bibinfo{number}{2} (\bibinfo{year}{2002}), \bibinfo{pages}{212--227}.
\newblock


\bibitem[Bavelas et~al\mbox{.}(1986)]%
        {bavelas1986show}
\bibfield{author}{\bibinfo{person}{Janet~B Bavelas}, \bibinfo{person}{Alex Black}, \bibinfo{person}{Charles~R Lemery}, {and} \bibinfo{person}{Jennifer Mullett}.} \bibinfo{year}{1986}\natexlab{}.
\newblock \showarticletitle{" I show how you feel": Motor mimicry as a communicative act.}
\newblock \bibinfo{journal}{\emph{Journal of personality and social psychology}} \bibinfo{volume}{50}, \bibinfo{number}{2} (\bibinfo{year}{1986}), \bibinfo{pages}{322}.
\newblock


\bibitem[Ben-Youssef et~al\mbox{.}(2019)]%
        {ben2019early}
\bibfield{author}{\bibinfo{person}{Atef Ben-Youssef}, \bibinfo{person}{Chlo{\'e} Clavel}, {and} \bibinfo{person}{Slim Essid}.} \bibinfo{year}{2019}\natexlab{}.
\newblock \showarticletitle{Early detection of user engagement breakdown in spontaneous human-humanoid interaction}.
\newblock \bibinfo{journal}{\emph{IEEE Transactions on Affective Computing}} \bibinfo{volume}{12}, \bibinfo{number}{3} (\bibinfo{year}{2019}), \bibinfo{pages}{776--787}.
\newblock


\bibitem[Ben-Youssef et~al\mbox{.}(2017)]%
        {Ben-Youssef:2017:UE-HRI_Dataset}
\bibfield{author}{\bibinfo{person}{Atef Ben-Youssef}, \bibinfo{person}{Chlo{\'e} Clavel}, \bibinfo{person}{Slim Essid}, \bibinfo{person}{Miriam Bilac}, \bibinfo{person}{Marine Chamoux}, {and} \bibinfo{person}{Angelica Lim}.} \bibinfo{year}{2017}\natexlab{}.
\newblock \showarticletitle{UE-HRI: A New Dataset for the Study of User Engagement in Spontaneous Human-robot Interactions}. In \bibinfo{booktitle}{\emph{Proceedings of the 19th ACM International Conference on Multimodal Interaction}} (Glasgow, UK) \emph{(\bibinfo{series}{ICMI 2017})}. \bibinfo{publisher}{ACM}, \bibinfo{address}{New York, NY, USA}, \bibinfo{pages}{464--472}.
\newblock
\showISBNx{978-1-4503-5543-8}
\urldef\tempurl%
\url{https://doi.org/10.1145/3136755.3136814}
\showDOI{\tempurl}


\bibitem[Bethel and Murphy(2007)]%
        {bethel2007survey}
\bibfield{author}{\bibinfo{person}{Cindy~L Bethel} {and} \bibinfo{person}{Robin~R Murphy}.} \bibinfo{year}{2007}\natexlab{}.
\newblock \showarticletitle{Survey of non-facial/non-verbal affective expressions for appearance-constrained robots}.
\newblock \bibinfo{journal}{\emph{IEEE Transactions on Systems, Man, and Cybernetics, Part C (Applications and Reviews)}} \bibinfo{volume}{38}, \bibinfo{number}{1} (\bibinfo{year}{2007}), \bibinfo{pages}{83--92}.
\newblock


\bibitem[Bickmore and Cassell(2001)]%
        {2001bickmore}
\bibfield{author}{\bibinfo{person}{Timothy Bickmore} {and} \bibinfo{person}{Justine Cassell}.} \bibinfo{year}{2001}\natexlab{}.
\newblock \showarticletitle{Relational Agents: A Model and Implementation of Building User Trust}. In \bibinfo{booktitle}{\emph{Proceedings of the SIGCHI Conference on Human Factors in Computing Systems}} (Seattle, Washington, USA) \emph{(\bibinfo{series}{CHI '01})}. \bibinfo{publisher}{Association for Computing Machinery}, \bibinfo{address}{New York, NY, USA}, \bibinfo{pages}{396–403}.
\newblock
\showISBNx{1581133278}
\urldef\tempurl%
\url{https://doi.org/10.1145/365024.365304}
\showDOI{\tempurl}


\bibitem[Blair(2003)]%
        {blair2003facial}
\bibfield{author}{\bibinfo{person}{RJR Blair}.} \bibinfo{year}{2003}\natexlab{}.
\newblock \showarticletitle{Facial expressions, their communicatory functions and neuro--cognitive substrates}.
\newblock \bibinfo{journal}{\emph{Philosophical Transactions of the Royal Society of London. Series B: Biological Sciences}} \bibinfo{volume}{358}, \bibinfo{number}{1431} (\bibinfo{year}{2003}), \bibinfo{pages}{561--572}.
\newblock


\bibitem[Bonchek-Dokow and Kaminka(2014)]%
        {bonchek2014towards}
\bibfield{author}{\bibinfo{person}{Elisheva Bonchek-Dokow} {and} \bibinfo{person}{Gal~A Kaminka}.} \bibinfo{year}{2014}\natexlab{}.
\newblock \showarticletitle{Towards computational models of intention detection and intention prediction}.
\newblock \bibinfo{journal}{\emph{Cognitive Systems Research}}  \bibinfo{volume}{28} (\bibinfo{year}{2014}), \bibinfo{pages}{44--79}.
\newblock


\bibitem[Breazeal et~al\mbox{.}(2005)]%
        {breazeal2005effects}
\bibfield{author}{\bibinfo{person}{C. Breazeal}, \bibinfo{person}{C.D. Kidd}, \bibinfo{person}{A.L. Thomaz}, \bibinfo{person}{G. Hoffman}, {and} \bibinfo{person}{M. Berlin}.} \bibinfo{year}{2005}\natexlab{}.
\newblock \showarticletitle{Effects of nonverbal communication on efficiency and robustness in human-robot teamwork}. In \bibinfo{booktitle}{\emph{2005 IEEE/RSJ International Conference on Intelligent Robots and Systems}}. \bibinfo{publisher}{IEEE}, \bibinfo{address}{New York, NY, USA}, \bibinfo{pages}{708--713}.
\newblock
\urldef\tempurl%
\url{https://doi.org/10.1109/IROS.2005.1545011}
\showDOI{\tempurl}


\bibitem[Bremers et~al\mbox{.}(2023)]%
        {bremers2023bystander}
\bibfield{author}{\bibinfo{person}{Alexandra Bremers}, \bibinfo{person}{Maria~Teresa Parreira}, \bibinfo{person}{Xuanyu Fang}, \bibinfo{person}{Natalie Friedman}, \bibinfo{person}{Adolfo Ramirez-Aristizabal}, \bibinfo{person}{Alexandria Pabst}, \bibinfo{person}{Mirjana Spasojevic}, \bibinfo{person}{Michael Kuniavsky}, {and} \bibinfo{person}{Wendy Ju}.} \bibinfo{year}{2023}\natexlab{}.
\newblock \showarticletitle{The Bystander Affect Detection (BAD) Dataset for Failure Detection in HRI}. In \bibinfo{booktitle}{\emph{2023 IEEE/RSJ International Conference on Intelligent Robots and Systems (IROS)}}. \bibinfo{pages}{11443--11450}.
\newblock
\urldef\tempurl%
\url{https://doi.org/10.1109/IROS55552.2023.10342442}
\showDOI{\tempurl}


\bibitem[Brodersen et~al\mbox{.}(2010)]%
        {accbalance}
\bibfield{author}{\bibinfo{person}{Kay~Henning Brodersen}, \bibinfo{person}{Cheng~Soon Ong}, \bibinfo{person}{Klaas~Enno Stephan}, {and} \bibinfo{person}{Joachim~M. Buhmann}.} \bibinfo{year}{2010}\natexlab{}.
\newblock \showarticletitle{The Balanced Accuracy and Its Posterior Distribution}. In \bibinfo{booktitle}{\emph{2010 20th International Conference on Pattern Recognition}}. \bibinfo{pages}{3121--3124}.
\newblock
\urldef\tempurl%
\url{https://doi.org/10.1109/ICPR.2010.764}
\showDOI{\tempurl}


\bibitem[Broekens(2007)]%
        {broekens2007emotion}
\bibfield{author}{\bibinfo{person}{Joost Broekens}.} \bibinfo{year}{2007}\natexlab{}.
\newblock \showarticletitle{Emotion and reinforcement: affective facial expressions facilitate robot learning}.
\newblock In \bibinfo{booktitle}{\emph{Artifical intelligence for human computing}}. \bibinfo{publisher}{Springer}, \bibinfo{address}{Cham}, \bibinfo{pages}{113--132}.
\newblock


\bibitem[Brooks(2017)]%
        {brooks2017human}
\bibfield{author}{\bibinfo{person}{Daniel~J Brooks}.} \bibinfo{year}{2017}\natexlab{}.
\newblock \emph{\bibinfo{title}{A human-centric approach to autonomous robot failures}}.
\newblock \bibinfo{thesistype}{Ph.\,D. Dissertation}. \bibinfo{school}{University of Massachusetts Lowell}.
\newblock


\bibitem[Bu et~al\mbox{.}(2024)]%
        {bu2024field}
\bibfield{author}{\bibinfo{person}{Fanjun Bu}, \bibinfo{person}{Alexandra Bremers}, \bibinfo{person}{Mark Colley}, {and} \bibinfo{person}{Wendy Ju}.} \bibinfo{year}{2024}\natexlab{}.
\newblock \showarticletitle{Field Notes on Deploying Research Robots in Public Spaces}. In \bibinfo{booktitle}{\emph{Extended Abstracts of the 2024 ACM International Conference on Human Computer Interaction (CHI)}}.
\newblock


\bibitem[Busso et~al\mbox{.}(2008)]%
        {busso2008iemocap}
\bibfield{author}{\bibinfo{person}{Carlos Busso}, \bibinfo{person}{Murtaza Bulut}, \bibinfo{person}{Chi-Chun Lee}, \bibinfo{person}{Abe Kazemzadeh}, \bibinfo{person}{Emily Mower}, \bibinfo{person}{Samuel Kim}, \bibinfo{person}{Jeannette~N Chang}, \bibinfo{person}{Sungbok Lee}, {and} \bibinfo{person}{Shrikanth~S Narayanan}.} \bibinfo{year}{2008}\natexlab{}.
\newblock \showarticletitle{IEMOCAP: Interactive emotional dyadic motion capture database}.
\newblock \bibinfo{journal}{\emph{Language resources and evaluation}} \bibinfo{volume}{42}, \bibinfo{number}{4} (\bibinfo{year}{2008}), \bibinfo{pages}{335--359}.
\newblock


\bibitem[Callon(1986)]%
        {callon1986sociology}
\bibfield{author}{\bibinfo{person}{Michel Callon}.} \bibinfo{year}{1986}\natexlab{}.
\newblock \showarticletitle{The sociology of an actor-network: The case of the electric vehicle}.
\newblock In \bibinfo{booktitle}{\emph{Mapping the dynamics of science and technology}}. \bibinfo{publisher}{Springer}, \bibinfo{address}{Cham}, \bibinfo{pages}{19--34}.
\newblock


\bibitem[Candon(2024)]%
        {candon2024leveraging}
\bibfield{author}{\bibinfo{person}{Kate Candon}.} \bibinfo{year}{2024}\natexlab{}.
\newblock \showarticletitle{Leveraging Implicit Human Feedback to Better Learn from Explicit Human Feedback in Human-Robot Interactions}. In \bibinfo{booktitle}{\emph{Companion of the 2024 ACM/IEEE International Conference on Human-Robot Interaction}} \emph{(\bibinfo{series}{HRI '24})}. \bibinfo{publisher}{Association for Computing Machinery}, \bibinfo{address}{New York, NY, USA}, \bibinfo{pages}{100–102}.
\newblock
\showISBNx{9798400703232}
\urldef\tempurl%
\url{https://doi.org/10.1145/3610978.3638368}
\showDOI{\tempurl}


\bibitem[Candon et~al\mbox{.}(2023a)]%
        {candon2023nonverbal}
\bibfield{author}{\bibinfo{person}{Kate Candon}, \bibinfo{person}{Jesse Chen}, \bibinfo{person}{Yoony Kim}, \bibinfo{person}{Zoe Hsu}, \bibinfo{person}{Nathan Tsoi}, {and} \bibinfo{person}{Marynel V\'{a}zquez}.} \bibinfo{year}{2023}\natexlab{a}.
\newblock \showarticletitle{Nonverbal Human Signals Can Help Autonomous Agents Infer Human Preferences for Their Behavior}. In \bibinfo{booktitle}{\emph{Proceedings of the 2023 International Conference on Autonomous Agents and Multiagent Systems}} (London, United Kingdom) \emph{(\bibinfo{series}{AAMAS '23})}. \bibinfo{publisher}{International Foundation for Autonomous Agents and Multiagent Systems}, \bibinfo{address}{Richland, SC}, \bibinfo{pages}{307–316}.
\newblock
\showISBNx{9781450394321}


\bibitem[Candon et~al\mbox{.}(2023b)]%
        {candon2023nonverbalhuman}
\bibfield{author}{\bibinfo{person}{Kate Candon}, \bibinfo{person}{Jesse Chen}, \bibinfo{person}{Yoony Kim}, \bibinfo{person}{Zoe Hsu}, \bibinfo{person}{Nathan Tsoi}, {and} \bibinfo{person}{Marynel V\'{a}zquez}.} \bibinfo{year}{2023}\natexlab{b}.
\newblock \showarticletitle{Nonverbal Human Signals Can Help Autonomous Agents Infer Human Preferences for Their Behavior}. In \bibinfo{booktitle}{\emph{Proceedings of the 2023 International Conference on Autonomous Agents and Multiagent Systems}} (London, United Kingdom) \emph{(\bibinfo{series}{AAMAS '23})}. \bibinfo{publisher}{International Foundation for Autonomous Agents and Multiagent Systems}, \bibinfo{address}{Richland, SC}, \bibinfo{pages}{307–316}.
\newblock
\showISBNx{9781450394321}


\bibitem[Cao et~al\mbox{.}(2017)]%
        {cao2017survey}
\bibfield{author}{\bibinfo{person}{Hoang-Long Cao}, \bibinfo{person}{Pablo~G{\'o}mez Esteban}, \bibinfo{person}{Albert De~Beir}, \bibinfo{person}{Ramona Simut}, \bibinfo{person}{Greet van~de Perre}, \bibinfo{person}{Dirk Lefeber}, {and} \bibinfo{person}{Bram Vanderborght}.} \bibinfo{year}{2017}\natexlab{}.
\newblock \showarticletitle{A survey on behavior control architectures for social robots in healthcare interventions}.
\newblock \bibinfo{journal}{\emph{International Journal of Humanoid Robotics}} \bibinfo{volume}{14}, \bibinfo{number}{04} (\bibinfo{year}{2017}), \bibinfo{pages}{1750021}.
\newblock


\bibitem[Carbonell et~al\mbox{.}(2021)]%
        {carbonell2021comparing}
\bibfield{author}{\bibinfo{person}{Marcos~Fern{\'a}ndez Carbonell}, \bibinfo{person}{Magnus Boman}, {and} \bibinfo{person}{Petri Laukka}.} \bibinfo{year}{2021}\natexlab{}.
\newblock \showarticletitle{Comparing supervised and unsupervised approaches to multimodal emotion recognition}.
\newblock \bibinfo{journal}{\emph{PeerJ Computer Science}}  \bibinfo{volume}{7} (\bibinfo{year}{2021}), \bibinfo{pages}{e804}.
\newblock


\bibitem[Carlson and Murphy(2005)]%
        {carlson2005ugvs}
\bibfield{author}{\bibinfo{person}{Jennifer Carlson} {and} \bibinfo{person}{Robin~R Murphy}.} \bibinfo{year}{2005}\natexlab{}.
\newblock \showarticletitle{How UGVs physically fail in the field}.
\newblock \bibinfo{journal}{\emph{IEEE Transactions on robotics}} \bibinfo{volume}{21}, \bibinfo{number}{3} (\bibinfo{year}{2005}), \bibinfo{pages}{423--437}.
\newblock


\bibitem[Carreira and Zisserman(2017)]%
        {2017carreirakinetics}
\bibfield{author}{\bibinfo{person}{Joao Carreira} {and} \bibinfo{person}{Andrew Zisserman}.} \bibinfo{year}{2017}\natexlab{}.
\newblock \bibinfo{title}{Quo Vadis, Action Recognition? A New Model and the Kinetics Dataset}.
\newblock
\newblock
\urldef\tempurl%
\url{https://doi.org/10.48550/ARXIV.1705.07750}
\showDOI{\tempurl}


\bibitem[Chung et~al\mbox{.}(2015)]%
        {chung2015cost}
\bibfield{author}{\bibinfo{person}{Yu-An Chung}, \bibinfo{person}{Hsuan-Tien Lin}, {and} \bibinfo{person}{Shao-Wen Yang}.} \bibinfo{year}{2015}\natexlab{}.
\newblock \showarticletitle{Cost-aware pre-training for multiclass cost-sensitive deep learning}.
\newblock \bibinfo{journal}{\emph{arXiv preprint arXiv:1511.09337}} (\bibinfo{year}{2015}).
\newblock


\bibitem[Cid et~al\mbox{.}(2013)]%
        {2013cid}
\bibfield{author}{\bibinfo{person}{Felipe Cid}, \bibinfo{person}{José Prado}, \bibinfo{person}{Pablo Bustos}, {and} \bibinfo{person}{Pedro Núñez}.} \bibinfo{year}{2013}\natexlab{}.
\newblock \showarticletitle{A Real Time and Robust Facial Expression Recognition and Imitation approach for Affective Human-Robot Interaction Using Gabor filtering}.
\newblock \bibinfo{journal}{\emph{IEEE International Conference on Intelligent Robots and Systems}}.
\newblock
\urldef\tempurl%
\url{https://doi.org/10.1109/IROS.2013.6696662}
\showDOI{\tempurl}


\bibitem[Clark and Brennan({[n.\,d.]})]%
        {1991clark}
\bibfield{author}{\bibinfo{person}{Herbert~H. Clark} {and} \bibinfo{person}{Susan~E. Brennan}.} \bibinfo{year}{[n.\,d.]}\natexlab{}.
\newblock \showarticletitle{Grounding in communication.}
\newblock \bibinfo{publisher}{American Psychological Association}, \bibinfo{pages}{127--149}.
\newblock
\urldef\tempurl%
\url{https://doi.org/10.1037/10096-006}
\showDOI{\tempurl}


\bibitem[Cohen(1960)]%
        {cohenskappa}
\bibfield{author}{\bibinfo{person}{Jacob Cohen}.} \bibinfo{year}{1960}\natexlab{}.
\newblock \showarticletitle{A Coefficient of Agreement for Nominal Scales}.
\newblock \bibinfo{journal}{\emph{Educational and Psychological Measurement}} \bibinfo{volume}{20}, \bibinfo{number}{1} (\bibinfo{year}{1960}), \bibinfo{pages}{37--46}.
\newblock
\urldef\tempurl%
\url{https://doi.org/10.1177/001316446002000104}
\showDOI{\tempurl}
\showeprint{https://doi.org/10.1177/001316446002000104}


\bibitem[Cohen et~al\mbox{.}(2017)]%
        {cohen2017influence}
\bibfield{author}{\bibinfo{person}{Laura Cohen}, \bibinfo{person}{Mahdi Khoramshahi}, \bibinfo{person}{Robin~N Salesse}, \bibinfo{person}{Catherine Bortolon}, \bibinfo{person}{Piotr S{\l}owi{\'n}ski}, \bibinfo{person}{Chao Zhai}, \bibinfo{person}{Krasimira Tsaneva-Atanasova}, \bibinfo{person}{Mario Di~Bernardo}, \bibinfo{person}{Delphine Capdevielle}, \bibinfo{person}{Ludovic Marin}, {et~al\mbox{.}}} \bibinfo{year}{2017}\natexlab{}.
\newblock \showarticletitle{Influence of facial feedback during a cooperative human-robot task in schizophrenia}.
\newblock \bibinfo{journal}{\emph{Scientific reports}} \bibinfo{volume}{7}, \bibinfo{number}{1} (\bibinfo{year}{2017}), \bibinfo{pages}{1--10}.
\newblock


\bibitem[Cuadra et~al\mbox{.}(2021a)]%
        {cuadra2021look}
\bibfield{author}{\bibinfo{person}{Andrea Cuadra}, \bibinfo{person}{Hansol Lee}, \bibinfo{person}{Jason Cho}, {and} \bibinfo{person}{Wendy Ju}.} \bibinfo{year}{2021}\natexlab{a}.
\newblock \showarticletitle{Look at Me When I Talk to You: A Video Dataset to Enable Voice Assistants to Recognize Errors}.
\newblock \bibinfo{journal}{\emph{arXiv preprint arXiv:2104.07153}} (\bibinfo{year}{2021}).
\newblock


\bibitem[Cuadra et~al\mbox{.}(2021b)]%
        {cuadra2021my}
\bibfield{author}{\bibinfo{person}{Andrea Cuadra}, \bibinfo{person}{Shuran Li}, \bibinfo{person}{Hansol Lee}, \bibinfo{person}{Jason Cho}, {and} \bibinfo{person}{Wendy Ju}.} \bibinfo{year}{2021}\natexlab{b}.
\newblock \showarticletitle{My bad! repairing intelligent voice assistant errors improves interaction}.
\newblock \bibinfo{journal}{\emph{Proceedings of the ACM on Human-Computer Interaction}} \bibinfo{volume}{5}, \bibinfo{number}{CSCW1} (\bibinfo{year}{2021}), \bibinfo{pages}{1--24}.
\newblock


\bibitem[Das et~al\mbox{.}(2021)]%
        {das2021plsm}
\bibfield{author}{\bibinfo{person}{Dipayan Das}, \bibinfo{person}{Saumik Bhattacharya}, \bibinfo{person}{Umapada Pal}, {and} \bibinfo{person}{Sukalpa Chanda}.} \bibinfo{year}{2021}\natexlab{}.
\newblock \showarticletitle{PLSM: A Parallelized Liquid State Machine for Unintentional Action Detection}.
\newblock \bibinfo{journal}{\emph{arXiv preprint arXiv:2105.09909}} (\bibinfo{year}{2021}).
\newblock


\bibitem[de~Assis~Neto et~al\mbox{.}(2020)]%
        {de2020detecting}
\bibfield{author}{\bibinfo{person}{Silvano~Ramos de Assis~Neto}, \bibinfo{person}{Guto~Leoni Santos}, \bibinfo{person}{Elisson da Silva~Rocha}, \bibinfo{person}{Malika Bendechache}, \bibinfo{person}{Pierangelo Rosati}, \bibinfo{person}{Theo Lynn}, {and} \bibinfo{person}{Patricia Takako~Endo}.} \bibinfo{year}{2020}\natexlab{}.
\newblock \showarticletitle{Detecting human activities based on a multimodal sensor data set using a bidirectional long short-term memory model: a case study}.
\newblock In \bibinfo{booktitle}{\emph{Challenges and Trends in Multimodal Fall Detection for Healthcare}}. \bibinfo{publisher}{Springer}, \bibinfo{address}{Cham}, \bibinfo{pages}{31--51}.
\newblock


\bibitem[Deng et~al\mbox{.}(2019)]%
        {Deng_2019}
\bibfield{author}{\bibinfo{person}{Eric Deng}, \bibinfo{person}{Bilge Mutlu}, {and} \bibinfo{person}{Maja~J Mataric}.} \bibinfo{year}{2019}\natexlab{}.
\newblock \showarticletitle{Embodiment in Socially Interactive Robots}.
\newblock \bibinfo{journal}{\emph{Foundations and Trends in Robotics}} \bibinfo{volume}{7}, \bibinfo{number}{4} (\bibinfo{year}{2019}), \bibinfo{pages}{251--356}.
\newblock
\urldef\tempurl%
\url{https://doi.org/10.1561/2300000056}
\showDOI{\tempurl}


\bibitem[Dhall et~al\mbox{.}(2012)]%
        {dhall2012collecting}
\bibfield{author}{\bibinfo{person}{Abhinav Dhall}, \bibinfo{person}{Roland Goecke}, \bibinfo{person}{Simon Lucey}, {and} \bibinfo{person}{Tom Gedeon}.} \bibinfo{year}{2012}\natexlab{}.
\newblock \showarticletitle{Collecting large, richly annotated facial-expression databases from movies}.
\newblock \bibinfo{journal}{\emph{IEEE multimedia}} \bibinfo{volume}{19}, \bibinfo{number}{03} (\bibinfo{year}{2012}), \bibinfo{pages}{34--41}.
\newblock


\bibitem[Edinger and Patterson(1983)]%
        {edinger1983nonverbal}
\bibfield{author}{\bibinfo{person}{Joyce~A Edinger} {and} \bibinfo{person}{Miles~L Patterson}.} \bibinfo{year}{1983}\natexlab{}.
\newblock \showarticletitle{Nonverbal involvement and social control.}
\newblock \bibinfo{journal}{\emph{Psychological bulletin}} \bibinfo{volume}{93}, \bibinfo{number}{1} (\bibinfo{year}{1983}), \bibinfo{pages}{30}.
\newblock


\bibitem[Ekman(1992)]%
        {ekman1992there}
\bibfield{author}{\bibinfo{person}{Paul Ekman}.} \bibinfo{year}{1992}\natexlab{}.
\newblock \showarticletitle{Are there basic emotions?}
\newblock  (\bibinfo{year}{1992}).
\newblock


\bibitem[Ekman and Friesen(1978)]%
        {ekman1978facial}
\bibfield{author}{\bibinfo{person}{Paul Ekman} {and} \bibinfo{person}{Wallace~V Friesen}.} \bibinfo{year}{1978}\natexlab{}.
\newblock \showarticletitle{Facial action coding system}.
\newblock \bibinfo{journal}{\emph{Environmental Psychology \& Nonverbal Behavior}} (\bibinfo{year}{1978}).
\newblock


\bibitem[Epstein et~al\mbox{.}(2020)]%
        {epstein2020oops}
\bibfield{author}{\bibinfo{person}{Dave Epstein}, \bibinfo{person}{Boyuan Chen}, {and} \bibinfo{person}{Carl Vondrick}.} \bibinfo{year}{2020}\natexlab{}.
\newblock \showarticletitle{Oops! predicting unintentional action in video}. In \bibinfo{booktitle}{\emph{Proceedings of the IEEE/CVF conference on computer vision and pattern recognition}}. \bibinfo{publisher}{IEEE}, \bibinfo{address}{New York, NY, USA}, \bibinfo{pages}{919--929}.
\newblock


\bibitem[Epstein and Vondrick(2021)]%
        {epstein2021learning}
\bibfield{author}{\bibinfo{person}{Dave Epstein} {and} \bibinfo{person}{Carl Vondrick}.} \bibinfo{year}{2021}\natexlab{}.
\newblock \showarticletitle{Learning goals from failure}. In \bibinfo{booktitle}{\emph{Proceedings of the IEEE/CVF Conference on Computer Vision and Pattern Recognition}}. \bibinfo{publisher}{IEEE}, \bibinfo{address}{New York, NY, USA}, \bibinfo{pages}{11194--11204}.
\newblock


\bibitem[Erg{\"u}l(2021)]%
        {ergul2021case}
\bibfield{author}{\bibinfo{person}{Hilal Erg{\"u}l}.} \bibinfo{year}{2021}\natexlab{}.
\newblock \showarticletitle{The case for smiling? Nonverbal behavior and oral corrective feedback}.
\newblock \bibinfo{journal}{\emph{Journal of Psycholinguistic Research}} (\bibinfo{year}{2021}), \bibinfo{pages}{1--16}.
\newblock


\bibitem[Fan et~al\mbox{.}(2020)]%
        {fan2020facial}
\bibfield{author}{\bibinfo{person}{Yingruo Fan}, \bibinfo{person}{Victor~OK Li}, {and} \bibinfo{person}{Jacqueline~CK Lam}.} \bibinfo{year}{2020}\natexlab{}.
\newblock \showarticletitle{Facial expression recognition with deeply-supervised attention network}.
\newblock \bibinfo{journal}{\emph{IEEE transactions on affective computing}} \bibinfo{volume}{13}, \bibinfo{number}{2} (\bibinfo{year}{2020}), \bibinfo{pages}{1057--1071}.
\newblock


\bibitem[Feine et~al\mbox{.}(2019)]%
        {feine2019taxonomy}
\bibfield{author}{\bibinfo{person}{Jasper Feine}, \bibinfo{person}{Ulrich Gnewuch}, \bibinfo{person}{Stefan Morana}, {and} \bibinfo{person}{Alexander Maedche}.} \bibinfo{year}{2019}\natexlab{}.
\newblock \showarticletitle{A taxonomy of social cues for conversational agents}.
\newblock \bibinfo{journal}{\emph{International Journal of Human-Computer Studies}}  \bibinfo{volume}{132} (\bibinfo{year}{2019}), \bibinfo{pages}{138--161}.
\newblock


\bibitem[Fiore et~al\mbox{.}(2013)]%
        {fiore2013toward}
\bibfield{author}{\bibinfo{person}{Stephen~M Fiore}, \bibinfo{person}{Travis~J Wiltshire}, \bibinfo{person}{Emilio~JC Lobato}, \bibinfo{person}{Florian~G Jentsch}, \bibinfo{person}{Wesley~H Huang}, {and} \bibinfo{person}{Benjamin Axelrod}.} \bibinfo{year}{2013}\natexlab{}.
\newblock \showarticletitle{Toward understanding social cues and signals in human--robot interaction: effects of robot gaze and proxemic behavior}.
\newblock \bibinfo{journal}{\emph{Frontiers in psychology}}  \bibinfo{volume}{4} (\bibinfo{year}{2013}), \bibinfo{pages}{859}.
\newblock


\bibitem[Frederiksen et~al\mbox{.}(2022)]%
        {frederiksen2022robot}
\bibfield{author}{\bibinfo{person}{Morten~Roed Frederiksen}, \bibinfo{person}{Katrin Fischer}, {and} \bibinfo{person}{Maja Matari{\'c}}.} \bibinfo{year}{2022}\natexlab{}.
\newblock \showarticletitle{Robot Vulnerability and the Elicitation of User Empathy}. In \bibinfo{booktitle}{\emph{2022 31st IEEE International Conference on Robot and Human Interactive Communication (RO-MAN)}}. IEEE, \bibinfo{publisher}{IEEE}, \bibinfo{address}{New York, NY, USA}, \bibinfo{pages}{52--58}.
\newblock


\bibitem[Fridlund et~al\mbox{.}(1997)]%
        {fridlund1997facial}
\bibfield{author}{\bibinfo{person}{Alan~J Fridlund}, \bibinfo{person}{Carlos Crivelli}, \bibinfo{person}{Sergio Jarillo}, \bibinfo{person}{Jos{\'e}-Miguel Fern{\'a}ndez-Dols}, {and} \bibinfo{person}{James~A Russell}.} \bibinfo{year}{1997}\natexlab{}.
\newblock \showarticletitle{Facial expressions}.
\newblock \bibinfo{journal}{\emph{The psychology of facial expression}} (\bibinfo{year}{1997}), \bibinfo{pages}{103}.
\newblock


\bibitem[Ghazal et~al\mbox{.}(2019)]%
        {ghazal2019human}
\bibfield{author}{\bibinfo{person}{Sumaira Ghazal}, \bibinfo{person}{Umar~S Khan}, \bibinfo{person}{Muhammad Mubasher~Saleem}, \bibinfo{person}{Nasir Rashid}, {and} \bibinfo{person}{Javaid Iqbal}.} \bibinfo{year}{2019}\natexlab{}.
\newblock \showarticletitle{Human activity recognition using 2D skeleton data and supervised machine learning}.
\newblock \bibinfo{journal}{\emph{IET image processing}} \bibinfo{volume}{13}, \bibinfo{number}{13} (\bibinfo{year}{2019}), \bibinfo{pages}{2572--2578}.
\newblock


\bibitem[Giuliani et~al\mbox{.}(2015)]%
        {giuliani2015systematic}
\bibfield{author}{\bibinfo{person}{Manuel Giuliani}, \bibinfo{person}{Nicole Mirnig}, \bibinfo{person}{Gerald Stollnberger}, \bibinfo{person}{Susanne Stadler}, \bibinfo{person}{Roland Buchner}, {and} \bibinfo{person}{Manfred Tscheligi}.} \bibinfo{year}{2015}\natexlab{}.
\newblock \showarticletitle{Systematic analysis of video data from different human--robot interaction studies: a categorization of social signals during error situations}.
\newblock \bibinfo{journal}{\emph{Frontiers in psychology}}  \bibinfo{volume}{6} (\bibinfo{year}{2015}), \bibinfo{pages}{931}.
\newblock


\bibitem[Green et~al\mbox{.}(2022)]%
        {green2022s}
\bibfield{author}{\bibinfo{person}{Haley~N Green}, \bibinfo{person}{Md~Mofijul Islam}, \bibinfo{person}{Shahira Ali}, {and} \bibinfo{person}{Tariq Iqbal}.} \bibinfo{year}{2022}\natexlab{}.
\newblock \showarticletitle{Who's laughing nao? examining perceptions of failure in a humorous robot partner}. In \bibinfo{booktitle}{\emph{2022 17th ACM/IEEE International Conference on Human-Robot Interaction (HRI)}}. IEEE, \bibinfo{publisher}{IEEE}, \bibinfo{address}{New York, NY, USA}, \bibinfo{pages}{313--322}.
\newblock


\bibitem[Haase and Tepper(1972)]%
        {Haase1972}
\bibfield{author}{\bibinfo{person}{Richard~F. Haase} {and} \bibinfo{person}{Donald~T. Tepper}.} \bibinfo{year}{1972}\natexlab{}.
\newblock \showarticletitle{Nonverbal components of empathic communication.}
\newblock \bibinfo{journal}{\emph{Journal of Counseling Psychology}}  \bibinfo{volume}{19} (\bibinfo{date}{9} \bibinfo{year}{1972}), \bibinfo{pages}{417--424}.
\newblock
Issue 5.
\showISSN{1939-2168}
\urldef\tempurl%
\url{https://doi.org/10.1037/h0033188}
\showDOI{\tempurl}


\bibitem[Hayes et~al\mbox{.}(2016)]%
        {2016hayes}
\bibfield{author}{\bibinfo{person}{Cory~J. Hayes}, \bibinfo{person}{Maryam Moosaei}, {and} \bibinfo{person}{Laurel~D. Riek}.} \bibinfo{year}{2016}\natexlab{}.
\newblock \showarticletitle{Exploring implicit human responses to robot mistakes in a learning from demonstration task}. In \bibinfo{booktitle}{\emph{2016 25th IEEE International Symposium on Robot and Human Interactive Communication (RO-MAN)}}. \bibinfo{publisher}{IEEE}, \bibinfo{address}{New York, NY, USA}, \bibinfo{pages}{246--252}.
\newblock
\urldef\tempurl%
\url{https://doi.org/10.1109/ROMAN.2016.7745138}
\showDOI{\tempurl}


\bibitem[Hoffman and Ju(2014)]%
        {hoffman2014designing}
\bibfield{author}{\bibinfo{person}{Guy Hoffman} {and} \bibinfo{person}{Wendy Ju}.} \bibinfo{year}{2014}\natexlab{}.
\newblock \showarticletitle{Designing robots with movement in mind}.
\newblock \bibinfo{journal}{\emph{Journal of Human-Robot Interaction}} \bibinfo{volume}{3}, \bibinfo{number}{1} (\bibinfo{year}{2014}), \bibinfo{pages}{91--122}.
\newblock


\bibitem[Hollnagel(1991)]%
        {hollnagel1991phenotype}
\bibfield{author}{\bibinfo{person}{Erik Hollnagel}.} \bibinfo{year}{1991}\natexlab{}.
\newblock \showarticletitle{The phenotype of erroneous actions: Implications for HCI design}.
\newblock \bibinfo{journal}{\emph{Human-computer interaction and complex systems}} (\bibinfo{year}{1991}), \bibinfo{pages}{73--121}.
\newblock


\bibitem[Honig et~al\mbox{.}(2022)]%
        {honig2022taxonomy}
\bibfield{author}{\bibinfo{person}{Shanee Honig}, \bibinfo{person}{Alon Bartal}, \bibinfo{person}{Yisrael Parmet}, {and} \bibinfo{person}{Tal Oron-Gilad}.} \bibinfo{year}{2022}\natexlab{}.
\newblock \showarticletitle{Using Online Customer Reviews to Classify, Predict, and Learn About Domestic Robot Failures}.
\newblock \bibinfo{journal}{\emph{International Journal of Social Robotics}} (\bibinfo{date}{11} \bibinfo{year}{2022}), \bibinfo{pages}{1--26}.
\newblock
\urldef\tempurl%
\url{https://doi.org/10.1007/s12369-022-00929-3}
\showDOI{\tempurl}


\bibitem[Honig and Oron-Gilad(2018)]%
        {honig2018understanding}
\bibfield{author}{\bibinfo{person}{Shanee Honig} {and} \bibinfo{person}{Tal Oron-Gilad}.} \bibinfo{year}{2018}\natexlab{}.
\newblock \showarticletitle{Understanding and resolving failures in human-robot interaction: Literature review and model development}.
\newblock \bibinfo{journal}{\emph{Frontiers in psychology}}  \bibinfo{volume}{9} (\bibinfo{year}{2018}), \bibinfo{pages}{861}.
\newblock


\bibitem[Honig and Oron-Gilad(2021)]%
        {honig2021expect}
\bibfield{author}{\bibinfo{person}{Shanee Honig} {and} \bibinfo{person}{Tal Oron-Gilad}.} \bibinfo{year}{2021}\natexlab{}.
\newblock \showarticletitle{Expect the unexpected: Leveraging the human-robot ecosystem to handle unexpected robot failures}.
\newblock \bibinfo{journal}{\emph{Frontiers in Robotics and AI}}  \bibinfo{volume}{8} (\bibinfo{year}{2021}), \bibinfo{pages}{656385}.
\newblock


\bibitem[Huang and Mutlu(2016)]%
        {2016huanggaze}
\bibfield{author}{\bibinfo{person}{Chien-Ming Huang} {and} \bibinfo{person}{Bilge Mutlu}.} \bibinfo{year}{2016}\natexlab{}.
\newblock \showarticletitle{Anticipatory robot control for efficient human-robot collaboration}. In \bibinfo{booktitle}{\emph{2016 11th ACM/IEEE International Conference on Human-Robot Interaction (HRI)}}. \bibinfo{publisher}{IEEE}, \bibinfo{address}{New York, NY, USA}, \bibinfo{pages}{83--90}.
\newblock
\urldef\tempurl%
\url{https://doi.org/10.1109/HRI.2016.7451737}
\showDOI{\tempurl}


\bibitem[Hwang and Matsumoto(2015)]%
        {hwang2015evidence}
\bibfield{author}{\bibinfo{person}{Hyisung Hwang} {and} \bibinfo{person}{David Matsumoto}.} \bibinfo{year}{2015}\natexlab{}.
\newblock \showarticletitle{Evidence for the universality of facial expressions of emotion}.
\newblock In \bibinfo{booktitle}{\emph{Understanding facial expressions in communication}}. \bibinfo{publisher}{Springer}, \bibinfo{address}{Cham}, \bibinfo{pages}{41--56}.
\newblock


\bibitem[Hwang et~al\mbox{.}(2014)]%
        {hwang2014reward}
\bibfield{author}{\bibinfo{person}{Kao-Shing Hwang}, \bibinfo{person}{JL Ling}, \bibinfo{person}{Yu-Ying Chen}, {and} \bibinfo{person}{Wei-Han Wang}.} \bibinfo{year}{2014}\natexlab{}.
\newblock \showarticletitle{Reward shaping for reinforcement learning by emotion expressions}. In \bibinfo{booktitle}{\emph{2014 IEEE International Conference on Systems, Man, and Cybernetics (SMC)}}. IEEE, \bibinfo{publisher}{IEEE}, \bibinfo{address}{New York, NY, USA}, \bibinfo{pages}{1288--1293}.
\newblock


\bibitem[Jack et~al\mbox{.}(2012)]%
        {jack2012facial}
\bibfield{author}{\bibinfo{person}{Rachael~E Jack}, \bibinfo{person}{Oliver~GB Garrod}, \bibinfo{person}{Hui Yu}, \bibinfo{person}{Roberto Caldara}, {and} \bibinfo{person}{Philippe~G Schyns}.} \bibinfo{year}{2012}\natexlab{}.
\newblock \showarticletitle{Facial expressions of emotion are not culturally universal}.
\newblock \bibinfo{journal}{\emph{Proceedings of the National Academy of Sciences}} \bibinfo{volume}{109}, \bibinfo{number}{19} (\bibinfo{year}{2012}), \bibinfo{pages}{7241--7244}.
\newblock


\bibitem[Jenni et~al\mbox{.}(2020)]%
        {jenni2020video}
\bibfield{author}{\bibinfo{person}{Simon Jenni}, \bibinfo{person}{Givi Meishvili}, {and} \bibinfo{person}{Paolo Favaro}.} \bibinfo{year}{2020}\natexlab{}.
\newblock \showarticletitle{Video representation learning by recognizing temporal transformations}. In \bibinfo{booktitle}{\emph{European Conference on Computer Vision}}. Springer, \bibinfo{publisher}{Springer}, \bibinfo{address}{Cham}, \bibinfo{pages}{425--442}.
\newblock


\bibitem[Johanson et~al\mbox{.}(2021)]%
        {johanson2021improving}
\bibfield{author}{\bibinfo{person}{Deborah~L Johanson}, \bibinfo{person}{Ho~Seok Ahn}, {and} \bibinfo{person}{Elizabeth Broadbent}.} \bibinfo{year}{2021}\natexlab{}.
\newblock \showarticletitle{Improving interactions with healthcare robots: A review of communication behaviours in social and healthcare contexts}.
\newblock \bibinfo{journal}{\emph{International Journal of Social Robotics}} \bibinfo{volume}{13}, \bibinfo{number}{8} (\bibinfo{year}{2021}), \bibinfo{pages}{1835--1850}.
\newblock


\bibitem[Jones et~al\mbox{.}(2011)]%
        {jones2011behavioral}
\bibfield{author}{\bibinfo{person}{Rebecca~M Jones}, \bibinfo{person}{Leah~H Somerville}, \bibinfo{person}{Jian Li}, \bibinfo{person}{Erika~J Ruberry}, \bibinfo{person}{Victoria Libby}, \bibinfo{person}{Gary Glover}, \bibinfo{person}{Henning~U Voss}, \bibinfo{person}{Douglas~J Ballon}, {and} \bibinfo{person}{BJ Casey}.} \bibinfo{year}{2011}\natexlab{}.
\newblock \showarticletitle{Behavioral and neural properties of social reinforcement learning}.
\newblock \bibinfo{journal}{\emph{Journal of Neuroscience}} \bibinfo{volume}{31}, \bibinfo{number}{37} (\bibinfo{year}{2011}), \bibinfo{pages}{13039--13045}.
\newblock


\bibitem[Kay et~al\mbox{.}(2017)]%
        {kay2017kinetics}
\bibfield{author}{\bibinfo{person}{Will Kay}, \bibinfo{person}{Joao Carreira}, \bibinfo{person}{Karen Simonyan}, \bibinfo{person}{Brian Zhang}, \bibinfo{person}{Chloe Hillier}, \bibinfo{person}{Sudheendra Vijayanarasimhan}, \bibinfo{person}{Fabio Viola}, \bibinfo{person}{Tim Green}, \bibinfo{person}{Trevor Back}, \bibinfo{person}{Paul Natsev}, {et~al\mbox{.}}} \bibinfo{year}{2017}\natexlab{}.
\newblock \showarticletitle{The kinetics human action video dataset}.
\newblock \bibinfo{journal}{\emph{arXiv preprint arXiv:1705.06950}} (\bibinfo{year}{2017}).
\newblock


\bibitem[Keltner and Buswell(1997)]%
        {keltner1997embarrassment}
\bibfield{author}{\bibinfo{person}{Dacher Keltner} {and} \bibinfo{person}{Brenda~N Buswell}.} \bibinfo{year}{1997}\natexlab{}.
\newblock \showarticletitle{Embarrassment: its distinct form and appeasement functions.}
\newblock \bibinfo{journal}{\emph{Psychological bulletin}} \bibinfo{volume}{122}, \bibinfo{number}{3} (\bibinfo{year}{1997}), \bibinfo{pages}{250}.
\newblock


\bibitem[Knox and Stone(2009)]%
        {knox2009interactively}
\bibfield{author}{\bibinfo{person}{W~Bradley Knox} {and} \bibinfo{person}{Peter Stone}.} \bibinfo{year}{2009}\natexlab{}.
\newblock \showarticletitle{Interactively shaping agents via human reinforcement: The TAMER framework}. In \bibinfo{booktitle}{\emph{Proceedings of the fifth international conference on Knowledge capture}}. \bibinfo{pages}{9--16}.
\newblock


\bibitem[Knox and Stone(2012)]%
        {knox2012reinforcement}
\bibfield{author}{\bibinfo{person}{W~Bradley Knox} {and} \bibinfo{person}{Peter Stone}.} \bibinfo{year}{2012}\natexlab{}.
\newblock \showarticletitle{Reinforcement learning from human reward: Discounting in episodic tasks}. In \bibinfo{booktitle}{\emph{2012 IEEE RO-MAN: The 21st IEEE international symposium on robot and human interactive communication}}. IEEE, \bibinfo{publisher}{IEEE}, \bibinfo{address}{New York, NY, USA}, \bibinfo{pages}{878--885}.
\newblock


\bibitem[Knox et~al\mbox{.}(2013)]%
        {knox2013training}
\bibfield{author}{\bibinfo{person}{W~Bradley Knox}, \bibinfo{person}{Peter Stone}, {and} \bibinfo{person}{Cynthia Breazeal}.} \bibinfo{year}{2013}\natexlab{}.
\newblock \showarticletitle{Training a robot via human feedback: A case study}. In \bibinfo{booktitle}{\emph{International Conference on Social Robotics}}. Springer, \bibinfo{publisher}{Springer}, \bibinfo{address}{Cham}, \bibinfo{pages}{460--470}.
\newblock


\bibitem[Kong and Fu(2022)]%
        {kong2022human}
\bibfield{author}{\bibinfo{person}{Yu Kong} {and} \bibinfo{person}{Yun Fu}.} \bibinfo{year}{2022}\natexlab{}.
\newblock \showarticletitle{Human action recognition and prediction: A survey}.
\newblock \bibinfo{journal}{\emph{International Journal of Computer Vision}} \bibinfo{volume}{130}, \bibinfo{number}{5} (\bibinfo{year}{2022}), \bibinfo{pages}{1366--1401}.
\newblock


\bibitem[Kontogiorgos et~al\mbox{.}(2020a)]%
        {kontogiorgos2020behavioural}
\bibfield{author}{\bibinfo{person}{Dimosthenis Kontogiorgos}, \bibinfo{person}{Andre Pereira}, \bibinfo{person}{Boran Sahindal}, \bibinfo{person}{Sanne van Waveren}, {and} \bibinfo{person}{Joakim Gustafson}.} \bibinfo{year}{2020}\natexlab{a}.
\newblock \showarticletitle{Behavioural responses to robot conversational failures}. In \bibinfo{booktitle}{\emph{2020 15th ACM/IEEE International Conference on Human-Robot Interaction (HRI)}}. IEEE, \bibinfo{publisher}{IEEE}, \bibinfo{address}{New York, NY, USA}, \bibinfo{pages}{53--62}.
\newblock


\bibitem[Kontogiorgos et~al\mbox{.}(2021)]%
        {kontogiorgos2021systematic}
\bibfield{author}{\bibinfo{person}{Dimosthenis Kontogiorgos}, \bibinfo{person}{Minh Tran}, \bibinfo{person}{Joakim Gustafson}, {and} \bibinfo{person}{Mohammad Soleymani}.} \bibinfo{year}{2021}\natexlab{}.
\newblock \showarticletitle{A systematic cross-corpus analysis of human reactions to robot conversational failures}. In \bibinfo{booktitle}{\emph{Proceedings of the 2021 International Conference on Multimodal Interaction}}. \bibinfo{pages}{112--120}.
\newblock


\bibitem[Kontogiorgos et~al\mbox{.}(2020b)]%
        {2020kontogiorgosembodiment}
\bibfield{author}{\bibinfo{person}{Dimosthenis Kontogiorgos}, \bibinfo{person}{Sanne van Waveren}, \bibinfo{person}{Olle Wallberg}, \bibinfo{person}{Andre Pereira}, \bibinfo{person}{Iolanda Leite}, {and} \bibinfo{person}{Joakim Gustafson}.} \bibinfo{year}{2020}\natexlab{b}.
\newblock \showarticletitle{Embodiment Effects in Interactions with Failing Robots}. In \bibinfo{booktitle}{\emph{Proceedings of the 2020 CHI Conference on Human Factors in Computing Systems}} (Honolulu, HI, USA) \emph{(\bibinfo{series}{CHI '20})}. \bibinfo{publisher}{Association for Computing Machinery}, \bibinfo{address}{New York, NY, USA}, \bibinfo{pages}{1–14}.
\newblock
\showISBNx{9781450367080}
\urldef\tempurl%
\url{https://doi.org/10.1145/3313831.3376372}
\showDOI{\tempurl}


\bibitem[Kuehne et~al\mbox{.}(2011)]%
        {kuehne2011hmdb}
\bibfield{author}{\bibinfo{person}{Hildegard Kuehne}, \bibinfo{person}{Hueihan Jhuang}, \bibinfo{person}{Est{\'\i}baliz Garrote}, \bibinfo{person}{Tomaso Poggio}, {and} \bibinfo{person}{Thomas Serre}.} \bibinfo{year}{2011}\natexlab{}.
\newblock \showarticletitle{HMDB: a large video database for human motion recognition}. In \bibinfo{booktitle}{\emph{2011 International conference on computer vision}}. IEEE, \bibinfo{publisher}{IEEE}, \bibinfo{address}{New York, NY, USA}, \bibinfo{pages}{2556--2563}.
\newblock


\bibitem[Kulsoom et~al\mbox{.}(2022)]%
        {kulsoom2022review}
\bibfield{author}{\bibinfo{person}{Farzana Kulsoom}, \bibinfo{person}{Sanam Narejo}, \bibinfo{person}{Zahid Mehmood}, \bibinfo{person}{Hassan~Nazeer Chaudhry}, \bibinfo{person}{Ali~Kashif Bashir}, {et~al\mbox{.}}} \bibinfo{year}{2022}\natexlab{}.
\newblock \showarticletitle{A review of machine learning-based human activity recognition for diverse applications}.
\newblock \bibinfo{journal}{\emph{Neural Computing and Applications}} (\bibinfo{year}{2022}), \bibinfo{pages}{1--36}.
\newblock


\bibitem[Lane et~al\mbox{.}(2006)]%
        {lane2006applying}
\bibfield{author}{\bibinfo{person}{Rhonda Lane}, \bibinfo{person}{Neville~A Stanton}, {and} \bibinfo{person}{David Harrison}.} \bibinfo{year}{2006}\natexlab{}.
\newblock \showarticletitle{Applying hierarchical task analysis to medication administration errors}.
\newblock \bibinfo{journal}{\emph{Applied ergonomics}} \bibinfo{volume}{37}, \bibinfo{number}{5} (\bibinfo{year}{2006}), \bibinfo{pages}{669--679}.
\newblock


\bibitem[Laprie(1985)]%
        {laprie1985dependable}
\bibfield{author}{\bibinfo{person}{Jean-Claude Laprie}.} \bibinfo{year}{1985}\natexlab{}.
\newblock \showarticletitle{Dependable computing and fault-tolerance}.
\newblock \bibinfo{journal}{\emph{Digest of Papers FTCS-15}} \bibinfo{volume}{10}, \bibinfo{number}{2} (\bibinfo{year}{1985}), \bibinfo{pages}{124}.
\newblock


\bibitem[Latour(1996)]%
        {latour1996actor}
\bibfield{author}{\bibinfo{person}{Bruno Latour}.} \bibinfo{year}{1996}\natexlab{}.
\newblock \showarticletitle{On actor-network theory: A few clarifications}.
\newblock \bibinfo{journal}{\emph{Soziale welt}} (\bibinfo{year}{1996}), \bibinfo{pages}{369--381}.
\newblock


\bibitem[Latour(2007)]%
        {latour2007reassembling}
\bibfield{author}{\bibinfo{person}{Bruno Latour}.} \bibinfo{year}{2007}\natexlab{}.
\newblock \bibinfo{booktitle}{\emph{Reassembling the social: An introduction to actor-network-theory}}.
\newblock \bibinfo{publisher}{Oup Oxford}.
\newblock


\bibitem[Leathers(1976)]%
        {leathers1976nonverbal}
\bibfield{author}{\bibinfo{person}{Dale~G Leathers}.} \bibinfo{year}{1976}\natexlab{}.
\newblock \showarticletitle{Nonverbal communication systems.}
\newblock  (\bibinfo{year}{1976}).
\newblock


\bibitem[Lewis et~al\mbox{.}(2009)]%
        {lewis2009using}
\bibfield{author}{\bibinfo{person}{Michael Lewis}, \bibinfo{person}{Huadong Wang}, \bibinfo{person}{Prasanna Velagapudi}, \bibinfo{person}{Paul Scerri}, {and} \bibinfo{person}{Katia Sycara}.} \bibinfo{year}{2009}\natexlab{}.
\newblock \showarticletitle{Using humans as sensors in robotic search}. In \bibinfo{booktitle}{\emph{2009 12th International Conference on Information Fusion}}. IEEE, \bibinfo{publisher}{IEEE}, \bibinfo{address}{New York, NY, USA}, \bibinfo{pages}{1249--1256}.
\newblock


\bibitem[Li et~al\mbox{.}(2020)]%
        {li2020facial}
\bibfield{author}{\bibinfo{person}{Guangliang Li}, \bibinfo{person}{Hamdi Dibeklio{\u{g}}lu}, \bibinfo{person}{Shimon Whiteson}, {and} \bibinfo{person}{Hayley Hung}.} \bibinfo{year}{2020}\natexlab{}.
\newblock \showarticletitle{Facial feedback for reinforcement learning: a case study and offline analysis using the TAMER framework}.
\newblock \bibinfo{journal}{\emph{Autonomous Agents and Multi-Agent Systems}} \bibinfo{volume}{34}, \bibinfo{number}{1} (\bibinfo{year}{2020}), \bibinfo{pages}{1--29}.
\newblock


\bibitem[Li et~al\mbox{.}(2019)]%
        {li2019communicating}
\bibfield{author}{\bibinfo{person}{Jamy Li}, \bibinfo{person}{Andrea Cuadra}, \bibinfo{person}{Brian Mok}, \bibinfo{person}{Byron Reeves}, \bibinfo{person}{Jofish Kaye}, {and} \bibinfo{person}{Wendy Ju}.} \bibinfo{year}{2019}\natexlab{}.
\newblock \showarticletitle{Communicating dominance in a nonanthropomorphic robot using locomotion}.
\newblock \bibinfo{journal}{\emph{ACM Transactions on Human-Robot Interaction (THRI)}} \bibinfo{volume}{8}, \bibinfo{number}{1} (\bibinfo{year}{2019}), \bibinfo{pages}{1--14}.
\newblock


\bibitem[Li and Deng(2020)]%
        {li2020deep}
\bibfield{author}{\bibinfo{person}{Shan Li} {and} \bibinfo{person}{Weihong Deng}.} \bibinfo{year}{2020}\natexlab{}.
\newblock \showarticletitle{Deep facial expression recognition: A survey}.
\newblock \bibinfo{journal}{\emph{IEEE transactions on affective computing}} (\bibinfo{year}{2020}).
\newblock


\bibitem[Li et~al\mbox{.}(2017)]%
        {li2017reliable}
\bibfield{author}{\bibinfo{person}{Shan Li}, \bibinfo{person}{Weihong Deng}, {and} \bibinfo{person}{JunPing Du}.} \bibinfo{year}{2017}\natexlab{}.
\newblock \showarticletitle{Reliable crowdsourcing and deep locality-preserving learning for expression recognition in the wild}. In \bibinfo{booktitle}{\emph{Proceedings of the IEEE conference on computer vision and pattern recognition}}. \bibinfo{publisher}{IEEE}, \bibinfo{address}{New York, NY, USA}, \bibinfo{pages}{2852--2861}.
\newblock


\bibitem[Liau and Ryu(2021)]%
        {2021liau}
\bibfield{author}{\bibinfo{person}{Yee~Yeng Liau} {and} \bibinfo{person}{Kwangyeol Ryu}.} \bibinfo{year}{2021}\natexlab{}.
\newblock \showarticletitle{Status Recognition Using Pre-Trained YOLOv5 for Sustainable Human-Robot Collaboration (HRC) System in Mold Assembly}.
\newblock \bibinfo{journal}{\emph{Sustainability}} \bibinfo{volume}{13}, \bibinfo{number}{21} (\bibinfo{date}{Oct} \bibinfo{year}{2021}), \bibinfo{pages}{12044}.
\newblock
\showISSN{2071-1050}
\urldef\tempurl%
\url{https://doi.org/10.3390/su132112044}
\showDOI{\tempurl}


\bibitem[Lin et~al\mbox{.}(2020)]%
        {lin2020review}
\bibfield{author}{\bibinfo{person}{Jinying Lin}, \bibinfo{person}{Zhen Ma}, \bibinfo{person}{Randy Gomez}, \bibinfo{person}{Keisuke Nakamura}, \bibinfo{person}{Bo He}, {and} \bibinfo{person}{Guangliang Li}.} \bibinfo{year}{2020}\natexlab{}.
\newblock \showarticletitle{A review on interactive reinforcement learning from human social feedback}.
\newblock \bibinfo{journal}{\emph{IEEE Access}}  \bibinfo{volume}{8} (\bibinfo{year}{2020}), \bibinfo{pages}{120757--120765}.
\newblock


\bibitem[Lucey et~al\mbox{.}(2010)]%
        {lucey2010extended}
\bibfield{author}{\bibinfo{person}{Patrick Lucey}, \bibinfo{person}{Jeffrey~F Cohn}, \bibinfo{person}{Takeo Kanade}, \bibinfo{person}{Jason Saragih}, \bibinfo{person}{Zara Ambadar}, {and} \bibinfo{person}{Iain Matthews}.} \bibinfo{year}{2010}\natexlab{}.
\newblock \showarticletitle{The extended cohn-kanade dataset (ck+): A complete dataset for action unit and emotion-specified expression}. In \bibinfo{booktitle}{\emph{2010 ieee computer society conference on computer vision and pattern recognition-workshops}}. IEEE, \bibinfo{publisher}{IEEE}, \bibinfo{address}{New York, NY, USA}, \bibinfo{pages}{94--101}.
\newblock


\bibitem[Mandal(2014)]%
        {mandal2014nonverbal}
\bibfield{author}{\bibinfo{person}{Fatik~Baran Mandal}.} \bibinfo{year}{2014}\natexlab{}.
\newblock \showarticletitle{Nonverbal Communication in Humans}.
\newblock \bibinfo{journal}{\emph{Journal of Human Behavior in the Social Environment}} \bibinfo{volume}{24}, \bibinfo{number}{4} (\bibinfo{year}{2014}), \bibinfo{pages}{417--421}.
\newblock
\urldef\tempurl%
\url{https://doi.org/10.1080/10911359.2013.831288}
\showDOI{\tempurl}
\showeprint{https://doi.org/10.1080/10911359.2013.831288}


\bibitem[Marsden and Hollnagel(1996)]%
        {marsden1996human}
\bibfield{author}{\bibinfo{person}{Phil Marsden} {and} \bibinfo{person}{Erik Hollnagel}.} \bibinfo{year}{1996}\natexlab{}.
\newblock \showarticletitle{Human interaction with technology: The accidental user}.
\newblock \bibinfo{journal}{\emph{Acta Psychologica}} \bibinfo{volume}{91}, \bibinfo{number}{3} (\bibinfo{year}{1996}), \bibinfo{pages}{345--358}.
\newblock


\bibitem[Martinez and Valstar(2016)]%
        {martinez2016advances}
\bibfield{author}{\bibinfo{person}{Brais Martinez} {and} \bibinfo{person}{Michel~F Valstar}.} \bibinfo{year}{2016}\natexlab{}.
\newblock \showarticletitle{Advances, challenges, and opportunities in automatic facial expression recognition}.
\newblock \bibinfo{journal}{\emph{Advances in face detection and facial image analysis}} (\bibinfo{year}{2016}), \bibinfo{pages}{63--100}.
\newblock


\bibitem[Mart{\'\i}nez-Villase{\~n}or et~al\mbox{.}(2019)]%
        {martinez2019up}
\bibfield{author}{\bibinfo{person}{Lourdes Mart{\'\i}nez-Villase{\~n}or}, \bibinfo{person}{Hiram Ponce}, \bibinfo{person}{Jorge Brieva}, \bibinfo{person}{Ernesto Moya-Albor}, \bibinfo{person}{Jos{\'e} N{\'u}{\~n}ez-Mart{\'\i}nez}, {and} \bibinfo{person}{Carlos Pe{\~n}afort-Asturiano}.} \bibinfo{year}{2019}\natexlab{}.
\newblock \showarticletitle{UP-fall detection dataset: A multimodal approach}.
\newblock \bibinfo{journal}{\emph{Sensors}} \bibinfo{volume}{19}, \bibinfo{number}{9} (\bibinfo{year}{2019}), \bibinfo{pages}{1988}.
\newblock


\bibitem[Mavridis(2015)]%
        {mavridis2015review}
\bibfield{author}{\bibinfo{person}{Nikolaos Mavridis}.} \bibinfo{year}{2015}\natexlab{}.
\newblock \showarticletitle{A review of verbal and non-verbal human--robot interactive communication}.
\newblock \bibinfo{journal}{\emph{Robotics and Autonomous Systems}}  \bibinfo{volume}{63} (\bibinfo{year}{2015}), \bibinfo{pages}{22--35}.
\newblock


\bibitem[McColl and Nejat(2012)]%
        {2012mccoll}
\bibfield{author}{\bibinfo{person}{Derek McColl} {and} \bibinfo{person}{Goldie Nejat}.} \bibinfo{year}{2012}\natexlab{}.
\newblock \showarticletitle{Affect detection from body language during social HRI}. In \bibinfo{booktitle}{\emph{2012 IEEE RO-MAN: The 21st IEEE International Symposium on Robot and Human Interactive Communication}}. \bibinfo{publisher}{IEEE}, \bibinfo{address}{New York, NY, USA}, \bibinfo{pages}{1013--1018}.
\newblock
\urldef\tempurl%
\url{https://doi.org/10.1109/ROMAN.2012.6343882}
\showDOI{\tempurl}


\bibitem[McColl et~al\mbox{.}(2011)]%
        {mccoll2011human}
\bibfield{author}{\bibinfo{person}{Derek McColl}, \bibinfo{person}{Zhe Zhang}, {and} \bibinfo{person}{Goldie Nejat}.} \bibinfo{year}{2011}\natexlab{}.
\newblock \showarticletitle{Human body pose interpretation and classification for social human-robot interaction}.
\newblock \bibinfo{journal}{\emph{International Journal of Social Robotics}} \bibinfo{volume}{3}, \bibinfo{number}{3} (\bibinfo{year}{2011}), \bibinfo{pages}{313--332}.
\newblock


\bibitem[McDuff et~al\mbox{.}(2013)]%
        {mcduff2013affectiva}
\bibfield{author}{\bibinfo{person}{Daniel McDuff}, \bibinfo{person}{Rana Kaliouby}, \bibinfo{person}{Thibaud Senechal}, \bibinfo{person}{May Amr}, \bibinfo{person}{Jeffrey Cohn}, {and} \bibinfo{person}{Rosalind Picard}.} \bibinfo{year}{2013}\natexlab{}.
\newblock \showarticletitle{Affectiva-mit facial expression dataset (am-fed): Naturalistic and spontaneous facial expressions collected}. In \bibinfo{booktitle}{\emph{Proceedings of the IEEE conference on computer vision and pattern recognition workshops}}. \bibinfo{publisher}{IEEE}, \bibinfo{address}{New York, NY, USA}, \bibinfo{pages}{881--888}.
\newblock


\bibitem[Mehlmann et~al\mbox{.}(2014)]%
        {2014mehlmann}
\bibfield{author}{\bibinfo{person}{Gregor Mehlmann}, \bibinfo{person}{Markus H\"{a}ring}, \bibinfo{person}{Kathrin Janowski}, \bibinfo{person}{Tobias Baur}, \bibinfo{person}{Patrick Gebhard}, {and} \bibinfo{person}{Elisabeth Andr\'{e}}.} \bibinfo{year}{2014}\natexlab{}.
\newblock \showarticletitle{Exploring a Model of Gaze for Grounding in Multimodal HRI}. In \bibinfo{booktitle}{\emph{Proceedings of the 16th International Conference on Multimodal Interaction}} (Istanbul, Turkey) \emph{(\bibinfo{series}{ICMI '14})}. \bibinfo{publisher}{Association for Computing Machinery}, \bibinfo{address}{New York, NY, USA}, \bibinfo{pages}{247–254}.
\newblock
\showISBNx{9781450328852}
\urldef\tempurl%
\url{https://doi.org/10.1145/2663204.2663275}
\showDOI{\tempurl}


\bibitem[Mirnig et~al\mbox{.}(2015)]%
        {2015mirnig}
\bibfield{author}{\bibinfo{person}{Nicole Mirnig}, \bibinfo{person}{Manuel Giuliani}, \bibinfo{person}{Gerald Stollnberger}, \bibinfo{person}{Susanne Stadler}, \bibinfo{person}{Roland Buchner}, {and} \bibinfo{person}{Manfred Tscheligi}.} \bibinfo{year}{2015}\natexlab{}.
\newblock \showarticletitle{Impact of Robot Actions on Social Signals and Reaction Times in HRI Error Situations}. In \bibinfo{booktitle}{\emph{Social Robotics}}, \bibfield{editor}{\bibinfo{person}{Adriana Tapus}, \bibinfo{person}{Elisabeth Andr{\'e}}, \bibinfo{person}{Jean-Claude Martin}, \bibinfo{person}{Fran{\c{c}}ois Ferland}, {and} \bibinfo{person}{Mehdi Ammi}} (Eds.). \bibinfo{publisher}{Springer International Publishing}, \bibinfo{address}{Cham}, \bibinfo{pages}{461--471}.
\newblock
\showISBNx{978-3-319-25554-5}


\bibitem[Mok et~al\mbox{.}(2015)]%
        {mok2015performing}
\bibfield{author}{\bibinfo{person}{Brian Mok}, \bibinfo{person}{Stephen Yang}, \bibinfo{person}{David Sirkin}, {and} \bibinfo{person}{Wendy Ju}.} \bibinfo{year}{2015}\natexlab{}.
\newblock \showarticletitle{Performing Collaborative Tasks with Robotic Drawers}. In \bibinfo{booktitle}{\emph{Proceedings of the Tenth Annual ACM/IEEE International Conference on Human-Robot Interaction Extended Abstracts}}. \bibinfo{pages}{309--309}.
\newblock


\bibitem[Morales et~al\mbox{.}(2019)]%
        {2019morales}
\bibfield{author}{\bibinfo{person}{Cecilia~G. Morales}, \bibinfo{person}{Elizabeth~J. Carter}, \bibinfo{person}{Xiang~Zhi Tan}, {and} \bibinfo{person}{Aaron Steinfeld}.} \bibinfo{year}{2019}\natexlab{}.
\newblock \showarticletitle{Interaction Needs and Opportunities for Failing Robots}. In \bibinfo{booktitle}{\emph{Proceedings of the 2019 on Designing Interactive Systems Conference}} (San Diego, CA, USA) \emph{(\bibinfo{series}{DIS '19})}. \bibinfo{publisher}{Association for Computing Machinery}, \bibinfo{address}{New York, NY, USA}, \bibinfo{pages}{659–670}.
\newblock
\showISBNx{9781450358507}
\urldef\tempurl%
\url{https://doi.org/10.1145/3322276.3322345}
\showDOI{\tempurl}


\bibitem[Mukherjee et~al\mbox{.}(2022)]%
        {mukherjee2022survey}
\bibfield{author}{\bibinfo{person}{Debasmita Mukherjee}, \bibinfo{person}{Kashish Gupta}, \bibinfo{person}{Li~Hsin Chang}, {and} \bibinfo{person}{Homayoun Najjaran}.} \bibinfo{year}{2022}\natexlab{}.
\newblock \showarticletitle{A survey of robot learning strategies for human-robot collaboration in industrial settings}.
\newblock \bibinfo{journal}{\emph{Robotics and Computer-Integrated Manufacturing}}  \bibinfo{volume}{73} (\bibinfo{year}{2022}), \bibinfo{pages}{102231}.
\newblock


\bibitem[Nass et~al\mbox{.}(1994)]%
        {nass1994computers}
\bibfield{author}{\bibinfo{person}{Clifford Nass}, \bibinfo{person}{Jonathan Steuer}, {and} \bibinfo{person}{Ellen~R Tauber}.} \bibinfo{year}{1994}\natexlab{}.
\newblock \showarticletitle{Computers are social actors}. In \bibinfo{booktitle}{\emph{Proceedings of the SIGCHI conference on Human factors in computing systems}}. \bibinfo{pages}{72--78}.
\newblock


\bibitem[Nielsen et~al\mbox{.}(2023)]%
        {nielsen2023using}
\bibfield{author}{\bibinfo{person}{Sara Nielsen}, \bibinfo{person}{Mikael~B. Skov}, \bibinfo{person}{Karl~Damkj\ae{}r Hansen}, {and} \bibinfo{person}{Aleksandra Kaszowska}.} \bibinfo{year}{2023}\natexlab{}.
\newblock \showarticletitle{Using User-Generated YouTube Videos to Understand Unguided Interactions with Robots in Public Places}.
\newblock \bibinfo{journal}{\emph{J. Hum.-Robot Interact.}} \bibinfo{volume}{12}, \bibinfo{number}{1}, Article \bibinfo{articleno}{5} (\bibinfo{date}{feb} \bibinfo{year}{2023}), \bibinfo{numpages}{40}~pages.
\newblock
\urldef\tempurl%
\url{https://doi.org/10.1145/3550280}
\showDOI{\tempurl}


\bibitem[Nwakanma et~al\mbox{.}(2021)]%
        {nwakanma2021detection}
\bibfield{author}{\bibinfo{person}{Cosmas~Ifeanyi Nwakanma}, \bibinfo{person}{Fabliha~Bushra Islam}, \bibinfo{person}{Mareska~Pratiwi Maharani}, \bibinfo{person}{Jae-Min Lee}, {and} \bibinfo{person}{Dong-Seong Kim}.} \bibinfo{year}{2021}\natexlab{}.
\newblock \showarticletitle{Detection and classification of human activity for emergency response in smart factory shop floor}.
\newblock \bibinfo{journal}{\emph{Applied Sciences}} \bibinfo{volume}{11}, \bibinfo{number}{8} (\bibinfo{year}{2021}), \bibinfo{pages}{3662}.
\newblock


\bibitem[O'Hare et~al\mbox{.}(2004)]%
        {o2004demonstrating}
\bibfield{author}{\bibinfo{person}{Greg~MP O'Hare}, \bibinfo{person}{Rem Collier}, {and} \bibinfo{person}{Robert Ross}.} \bibinfo{year}{2004}\natexlab{}.
\newblock \showarticletitle{Demonstrating social error recovery with agentfactory}. In \bibinfo{booktitle}{\emph{3rd International Joint Conference on Autonomous Agents and Multi Agent Systems (AAMAS04), New York, USA, 19-23 July 2004}}. IEEE, \bibinfo{publisher}{IEEE}, \bibinfo{address}{New York, NY, USA}.
\newblock


\bibitem[Park and Kitayama(2014)]%
        {park2014interdependent}
\bibfield{author}{\bibinfo{person}{Jiyoung Park} {and} \bibinfo{person}{Shinobu Kitayama}.} \bibinfo{year}{2014}\natexlab{}.
\newblock \showarticletitle{Interdependent selves show face-induced facilitation of error processing: Cultural neuroscience of self-threat}.
\newblock \bibinfo{journal}{\emph{Social cognitive and affective neuroscience}} \bibinfo{volume}{9}, \bibinfo{number}{2} (\bibinfo{year}{2014}), \bibinfo{pages}{201--208}.
\newblock


\bibitem[Parreira et~al\mbox{.}(2023)]%
        {parreira2023badidea}
\bibfield{author}{\bibinfo{person}{Maria~Teresa Parreira}, \bibinfo{person}{Sukruth Gowdru~Lingaraju}, \bibinfo{person}{Adolfo Ramirez-Aristizabal}, \bibinfo{person}{Alexandra Bremers}, \bibinfo{person}{Manaswi Saha}, \bibinfo{person}{Michael Kuniavsky}, {and} \bibinfo{person}{Wendy Ju}.} \bibinfo{year}{2023}\natexlab{}.
\newblock \showarticletitle{Bad Idea? Exploring Anticipatory Human Reactions for Outcome Prediction}. In \bibinfo{booktitle}{\emph{NERC Northeast Robotics Colloquium}}.
\newblock


\bibitem[Parreira et~al\mbox{.}(2024)]%
        {parreira2024study}
\bibfield{author}{\bibinfo{person}{Maria~Teresa Parreira}, \bibinfo{person}{Sukruth~Gowdru Lingaraju}, \bibinfo{person}{Adolfo Ramirez-Aristizabal}, \bibinfo{person}{Manaswi Saha}, \bibinfo{person}{Michael Kuniavsky}, {and} \bibinfo{person}{Wendy Ju}.} \bibinfo{year}{2024}\natexlab{}.
\newblock \showarticletitle{A Study on Domain Generalization for Failure Detection through Human Reactions in HRI}.
\newblock \bibinfo{journal}{\emph{arXiv preprint arXiv:2403.06315}} (\bibinfo{year}{2024}).
\newblock


\bibitem[Parrott and Smith(1991)]%
        {parrott1991embarrassment}
\bibfield{author}{\bibinfo{person}{W~Gerrod Parrott} {and} \bibinfo{person}{Stefanie~F Smith}.} \bibinfo{year}{1991}\natexlab{}.
\newblock \showarticletitle{Embarrassment: Actual vs. typical cases, classical vs. prototypical representations}.
\newblock \bibinfo{journal}{\emph{Cognition \& Emotion}} \bibinfo{volume}{5}, \bibinfo{number}{5-6} (\bibinfo{year}{1991}), \bibinfo{pages}{467--488}.
\newblock


\bibitem[Phipps et~al\mbox{.}(2008)]%
        {phipps2008human}
\bibfield{author}{\bibinfo{person}{Denham Phipps}, \bibinfo{person}{George~H Meakin}, \bibinfo{person}{Paul~CW Beatty}, \bibinfo{person}{Chidozie Nsoedo}, {and} \bibinfo{person}{Dianne Parker}.} \bibinfo{year}{2008}\natexlab{}.
\newblock \showarticletitle{Human factors in anaesthetic practice: insights from a task analysis}.
\newblock \bibinfo{journal}{\emph{British journal of anaesthesia}} \bibinfo{volume}{100}, \bibinfo{number}{3} (\bibinfo{year}{2008}), \bibinfo{pages}{333--343}.
\newblock


\bibitem[Poria et~al\mbox{.}(2018)]%
        {poria2018meld}
\bibfield{author}{\bibinfo{person}{Soujanya Poria}, \bibinfo{person}{Devamanyu Hazarika}, \bibinfo{person}{Navonil Majumder}, \bibinfo{person}{Gautam Naik}, \bibinfo{person}{Erik Cambria}, {and} \bibinfo{person}{Rada Mihalcea}.} \bibinfo{year}{2018}\natexlab{}.
\newblock \showarticletitle{Meld: A multimodal multi-party dataset for emotion recognition in conversations}.
\newblock \bibinfo{journal}{\emph{arXiv preprint arXiv:1810.02508}} (\bibinfo{year}{2018}).
\newblock


\bibitem[{Qualitative Data Repository}({[n.\,d.]})]%
        {qdr}
\bibfield{author}{\bibinfo{person}{{Qualitative Data Repository}}.} \bibinfo{year}{[n.\,d.]}\natexlab{}.
\newblock \bibinfo{title}{{Qualitative Data Repository}}.
\newblock \bibinfo{howpublished}{\url{https://qdr.syr.edu}}.
\newblock
\newblock
\shownote{Accessed: March 19, 2024}.


\bibitem[Rasmussen(1982)]%
        {rasmussen1982human}
\bibfield{author}{\bibinfo{person}{Jens Rasmussen}.} \bibinfo{year}{1982}\natexlab{}.
\newblock \showarticletitle{Human errors. A taxonomy for describing human malfunction in industrial installations}.
\newblock \bibinfo{journal}{\emph{Journal of occupational accidents}} \bibinfo{volume}{4}, \bibinfo{number}{2-4} (\bibinfo{year}{1982}), \bibinfo{pages}{311--333}.
\newblock


\bibitem[Read et~al\mbox{.}(2021)]%
        {read2021state}
\bibfield{author}{\bibinfo{person}{Gemma~JM Read}, \bibinfo{person}{Steven Shorrock}, \bibinfo{person}{Guy~H Walker}, {and} \bibinfo{person}{Paul~M Salmon}.} \bibinfo{year}{2021}\natexlab{}.
\newblock \showarticletitle{State of science: Evolving perspectives on ‘human error’}.
\newblock \bibinfo{journal}{\emph{Ergonomics}} \bibinfo{volume}{64}, \bibinfo{number}{9} (\bibinfo{year}{2021}), \bibinfo{pages}{1091--1114}.
\newblock


\bibitem[Reig et~al\mbox{.}(2021)]%
        {reig2021flailing}
\bibfield{author}{\bibinfo{person}{Samantha Reig}, \bibinfo{person}{Elizabeth~J Carter}, \bibinfo{person}{Terrence Fong}, \bibinfo{person}{Jodi Forlizzi}, {and} \bibinfo{person}{Aaron Steinfeld}.} \bibinfo{year}{2021}\natexlab{}.
\newblock \showarticletitle{Flailing, hailing, prevailing: Perceptions of multi-robot failure recovery strategies}. In \bibinfo{booktitle}{\emph{Proceedings of the 2021 ACM/IEEE International Conference on Human-Robot Interaction}}. \bibinfo{pages}{158--167}.
\newblock


\bibitem[Richter et~al\mbox{.}(2016)]%
        {2016richter}
\bibfield{author}{\bibinfo{person}{Viktor Richter}, \bibinfo{person}{Birte Carlmeyer}, \bibinfo{person}{Florian Lier}, \bibinfo{person}{Sebastian Meyer~zu Borgsen}, \bibinfo{person}{David Schlangen}, \bibinfo{person}{Franz Kummert}, \bibinfo{person}{Sven Wachsmuth}, {and} \bibinfo{person}{Britta Wrede}.} \bibinfo{year}{2016}\natexlab{}.
\newblock \showarticletitle{Are You Talking to Me? Improving the Robustness of Dialogue Systems in a Multi Party HRI Scenario by Incorporating Gaze Direction and Lip Movement of Attendees}. In \bibinfo{booktitle}{\emph{Proceedings of the Fourth International Conference on Human Agent Interaction}} (Biopolis, Singapore) \emph{(\bibinfo{series}{HAI '16})}. \bibinfo{publisher}{Association for Computing Machinery}, \bibinfo{address}{New York, NY, USA}, \bibinfo{pages}{43–50}.
\newblock
\showISBNx{9781450345088}
\urldef\tempurl%
\url{https://doi.org/10.1145/2974804.2974823}
\showDOI{\tempurl}


\bibitem[Rouse and Rouse(1983)]%
        {rouse1983analysis}
\bibfield{author}{\bibinfo{person}{William~B Rouse} {and} \bibinfo{person}{Sandra~H Rouse}.} \bibinfo{year}{1983}\natexlab{}.
\newblock \showarticletitle{Analysis and classification of human error}.
\newblock \bibinfo{journal}{\emph{IEEE Transactions on Systems, Man, and Cybernetics}} \bibinfo{number}{4} (\bibinfo{year}{1983}), \bibinfo{pages}{539--549}.
\newblock


\bibitem[Sagha et~al\mbox{.}(2011)]%
        {sagha2011benchmarking}
\bibfield{author}{\bibinfo{person}{Hesam Sagha}, \bibinfo{person}{Sundara~Tejaswi Digumarti}, \bibinfo{person}{Jos{\'e} del~R Mill{\'a}n}, \bibinfo{person}{Ricardo Chavarriaga}, \bibinfo{person}{Alberto Calatroni}, \bibinfo{person}{Daniel Roggen}, {and} \bibinfo{person}{Gerhard Tr{\"o}ster}.} \bibinfo{year}{2011}\natexlab{}.
\newblock \showarticletitle{Benchmarking classification techniques using the Opportunity human activity dataset}. In \bibinfo{booktitle}{\emph{2011 IEEE International Conference on Systems, Man, and Cybernetics}}. IEEE, \bibinfo{publisher}{IEEE}, \bibinfo{address}{New York, NY, USA}, \bibinfo{pages}{36--40}.
\newblock


\bibitem[Salem et~al\mbox{.}(2015)]%
        {2015salem}
\bibfield{author}{\bibinfo{person}{Maha Salem}, \bibinfo{person}{Gabriella Lakatos}, \bibinfo{person}{Farshid Amirabdollahian}, {and} \bibinfo{person}{Kerstin Dautenhahn}.} \bibinfo{year}{2015}\natexlab{}.
\newblock \showarticletitle{Would You Trust a (Faulty) Robot? Effects of Error, Task Type and Personality on Human-Robot Cooperation and Trust}. In \bibinfo{booktitle}{\emph{Proceedings of the Tenth Annual ACM/IEEE International Conference on Human-Robot Interaction}} (Portland, Oregon, USA) \emph{(\bibinfo{series}{HRI '15})}. \bibinfo{publisher}{Association for Computing Machinery}, \bibinfo{address}{New York, NY, USA}, \bibinfo{pages}{141–148}.
\newblock
\showISBNx{9781450328838}
\urldef\tempurl%
\url{https://doi.org/10.1145/2696454.2696497}
\showDOI{\tempurl}


\bibitem[Salem et~al\mbox{.}(2014)]%
        {2014salem}
\bibfield{author}{\bibinfo{person}{Maha Salem}, \bibinfo{person}{Micheline Ziadee}, {and} \bibinfo{person}{Majd Sakr}.} \bibinfo{year}{2014}\natexlab{}.
\newblock \showarticletitle{Marhaba, How May i Help You? Effects of Politeness and Culture on Robot Acceptance and Anthropomorphization}. In \bibinfo{booktitle}{\emph{Proceedings of the 2014 ACM/IEEE International Conference on Human-Robot Interaction}} (Bielefeld, Germany) \emph{(\bibinfo{series}{HRI '14})}. \bibinfo{publisher}{Association for Computing Machinery}, \bibinfo{address}{New York, NY, USA}, \bibinfo{pages}{74–81}.
\newblock
\showISBNx{9781450326582}
\urldef\tempurl%
\url{https://doi.org/10.1145/2559636.2559683}
\showDOI{\tempurl}


\bibitem[Sanghvi et~al\mbox{.}(2011)]%
        {2011sanghvi}
\bibfield{author}{\bibinfo{person}{Jyotirmay Sanghvi}, \bibinfo{person}{Ginevra Castellano}, \bibinfo{person}{Iolanda Leite}, \bibinfo{person}{Andr\'{e} Pereira}, \bibinfo{person}{Peter~W. McOwan}, {and} \bibinfo{person}{Ana Paiva}.} \bibinfo{year}{2011}\natexlab{}.
\newblock \showarticletitle{Automatic Analysis of Affective Postures and Body Motion to Detect Engagement with a Game Companion}. In \bibinfo{booktitle}{\emph{Proceedings of the 6th International Conference on Human-Robot Interaction}} (Lausanne, Switzerland) \emph{(\bibinfo{series}{HRI '11})}. \bibinfo{publisher}{Association for Computing Machinery}, \bibinfo{address}{New York, NY, USA}, \bibinfo{pages}{305–312}.
\newblock
\showISBNx{9781450305617}
\urldef\tempurl%
\url{https://doi.org/10.1145/1957656.1957781}
\showDOI{\tempurl}


\bibitem[Saunderson and Nejat(2019)]%
        {Saunderson2019}
\bibfield{author}{\bibinfo{person}{Shane Saunderson} {and} \bibinfo{person}{Goldie Nejat}.} \bibinfo{year}{2019}\natexlab{}.
\newblock \showarticletitle{How Robots Influence Humans: A Survey of Nonverbal Communication in Social Human–Robot Interaction}.
\newblock \bibinfo{journal}{\emph{International Journal of Social Robotics}}  \bibinfo{volume}{11} (\bibinfo{date}{8} \bibinfo{year}{2019}), \bibinfo{pages}{575--608}.
\newblock
Issue 4.
\showISSN{1875-4791}
\urldef\tempurl%
\url{https://doi.org/10.1007/s12369-019-00523-0}
\showDOI{\tempurl}


\bibitem[Scheirer et~al\mbox{.}(2002)]%
        {scheirer2002frustrating}
\bibfield{author}{\bibinfo{person}{Jocelyn Scheirer}, \bibinfo{person}{Raul Fernandez}, \bibinfo{person}{Jonathan Klein}, {and} \bibinfo{person}{Rosalind~W Picard}.} \bibinfo{year}{2002}\natexlab{}.
\newblock \showarticletitle{Frustrating the user on purpose: a step toward building an affective computer}.
\newblock \bibinfo{journal}{\emph{Interacting with computers}} \bibinfo{volume}{14}, \bibinfo{number}{2} (\bibinfo{year}{2002}), \bibinfo{pages}{93--118}.
\newblock


\bibitem[Sellen(1994)]%
        {sellen1994detection}
\bibfield{author}{\bibinfo{person}{Abigail~J Sellen}.} \bibinfo{year}{1994}\natexlab{}.
\newblock \showarticletitle{Detection of everyday errors}.
\newblock \bibinfo{journal}{\emph{Applied Psychology}} \bibinfo{volume}{43}, \bibinfo{number}{4} (\bibinfo{year}{1994}), \bibinfo{pages}{475--498}.
\newblock


\bibitem[Shi et~al\mbox{.}(2021)]%
        {2021shi}
\bibfield{author}{\bibinfo{person}{Lei Shi}, \bibinfo{person}{Cosmin Copot}, {and} \bibinfo{person}{Steve Vanlanduit}.} \bibinfo{year}{2021}\natexlab{}.
\newblock \showarticletitle{GazeEMD: Detecting Visual Intention in Gaze-Based Human-Robot Interaction}.
\newblock \bibinfo{journal}{\emph{Robotics}} \bibinfo{volume}{10}, \bibinfo{number}{2} (\bibinfo{year}{2021}).
\newblock
\showISSN{2218-6581}
\urldef\tempurl%
\url{https://doi.org/10.3390/robotics10020068}
\showDOI{\tempurl}


\bibitem[Shi et~al\mbox{.}(2019)]%
        {shi2019automatic}
\bibfield{author}{\bibinfo{person}{Zheng Shi}, \bibinfo{person}{Ya Zhang}, \bibinfo{person}{Cunling Bian}, {and} \bibinfo{person}{Weigang Lu}.} \bibinfo{year}{2019}\natexlab{}.
\newblock \showarticletitle{Automatic academic confusion recognition in online learning based on facial expressions}. In \bibinfo{booktitle}{\emph{2019 14th International Conference on Computer Science \& Education (ICCSE)}}. IEEE, \bibinfo{publisher}{IEEE}, \bibinfo{address}{New York, NY, USA}, \bibinfo{pages}{528--532}.
\newblock


\bibitem[Short et~al\mbox{.}(2018)]%
        {2018short}
\bibfield{author}{\bibinfo{person}{Elaine~Schaertl Short}, \bibinfo{person}{Mai~Lee Chang}, {and} \bibinfo{person}{Andrea Thomaz}.} \bibinfo{year}{2018}\natexlab{}.
\newblock \showarticletitle{Detecting Contingency for HRI in Open-World Environments}. In \bibinfo{booktitle}{\emph{Proceedings of the 2018 ACM/IEEE International Conference on Human-Robot Interaction}} (Chicago, IL, USA) \emph{(\bibinfo{series}{HRI '18})}. \bibinfo{publisher}{Association for Computing Machinery}, \bibinfo{address}{New York, NY, USA}, \bibinfo{pages}{425–433}.
\newblock
\showISBNx{9781450349536}
\urldef\tempurl%
\url{https://doi.org/10.1145/3171221.3171271}
\showDOI{\tempurl}


\bibitem[Skantze(2005)]%
        {2005skantze}
\bibfield{author}{\bibinfo{person}{Gabriel Skantze}.} \bibinfo{year}{2005}\natexlab{}.
\newblock \showarticletitle{Exploring human error recovery strategies: Implications for spoken dialogue systems}.
\newblock \bibinfo{journal}{\emph{Speech Communication}} \bibinfo{volume}{45}, \bibinfo{number}{3} (\bibinfo{year}{2005}), \bibinfo{pages}{325--341}.
\newblock
\showISSN{0167-6393}
\urldef\tempurl%
\url{https://doi.org/10.1016/j.specom.2004.11.005}
\showDOI{\tempurl}
\newblock
\shownote{Special Issue on Error Handling in Spoken Dialogue Systems}.


\bibitem[Soomro et~al\mbox{.}(2012)]%
        {soomro2012ucf101}
\bibfield{author}{\bibinfo{person}{Khurram Soomro}, \bibinfo{person}{Amir~Roshan Zamir}, {and} \bibinfo{person}{Mubarak Shah}.} \bibinfo{year}{2012}\natexlab{}.
\newblock \showarticletitle{UCF101: A dataset of 101 human actions classes from videos in the wild}.
\newblock \bibinfo{journal}{\emph{arXiv preprint arXiv:1212.0402}} (\bibinfo{year}{2012}).
\newblock


\bibitem[Srinivasa et~al\mbox{.}(2017)]%
        {srinivasa2017analysis}
\bibfield{author}{\bibinfo{person}{KG Srinivasa}, \bibinfo{person}{Sriram Anupindi}, \bibinfo{person}{R Sharath}, {and} \bibinfo{person}{S~Krishna Chaitanya}.} \bibinfo{year}{2017}\natexlab{}.
\newblock \showarticletitle{Analysis of facial expressiveness captured in reaction to videos}. In \bibinfo{booktitle}{\emph{2017 IEEE 7th International Advance Computing Conference (IACC)}}. IEEE, \bibinfo{publisher}{IEEE}, \bibinfo{address}{New York, NY, USA}, \bibinfo{pages}{664--670}.
\newblock


\bibitem[Steinbauer(2012)]%
        {steinbauer2012survey}
\bibfield{author}{\bibinfo{person}{Gerald Steinbauer}.} \bibinfo{year}{2012}\natexlab{}.
\newblock \showarticletitle{A survey about faults of robots used in robocup}. In \bibinfo{booktitle}{\emph{Robot Soccer World Cup}}. Springer, \bibinfo{publisher}{Springer}, \bibinfo{address}{Cham}, \bibinfo{pages}{344--355}.
\newblock


\bibitem[Stiber(2022)]%
        {stiber2022effective}
\bibfield{author}{\bibinfo{person}{Maia Stiber}.} \bibinfo{year}{2022}\natexlab{}.
\newblock \showarticletitle{Effective Human-Robot Collaboration via Generalized Robot Error Management Using Natural Human Responses}. In \bibinfo{booktitle}{\emph{Proceedings of the 2022 International Conference on Multimodal Interaction}} (Bengaluru, India) \emph{(\bibinfo{series}{ICMI '22})}. \bibinfo{publisher}{Association for Computing Machinery}, \bibinfo{address}{New York, NY, USA}, \bibinfo{pages}{673–678}.
\newblock
\showISBNx{9781450393904}
\urldef\tempurl%
\url{https://doi.org/10.1145/3536221.3557028}
\showDOI{\tempurl}


\bibitem[Stiber and Huang(2021)]%
        {2020stiber}
\bibfield{author}{\bibinfo{person}{Maia Stiber} {and} \bibinfo{person}{Chien-Ming Huang}.} \bibinfo{year}{2021}\natexlab{}.
\newblock \showarticletitle{Not All Errors Are Created Equal: Exploring Human Responses to Robot Errors with Varying Severity}. In \bibinfo{booktitle}{\emph{Companion Publication of the 2020 International Conference on Multimodal Interaction}} (Virtual Event, Netherlands) \emph{(\bibinfo{series}{ICMI '20 Companion})}. \bibinfo{publisher}{Association for Computing Machinery}, \bibinfo{address}{New York, NY, USA}, \bibinfo{pages}{97–101}.
\newblock
\showISBNx{9781450380027}
\urldef\tempurl%
\url{https://doi.org/10.1145/3395035.3425245}
\showDOI{\tempurl}


\bibitem[Stiber et~al\mbox{.}(2022)]%
        {2022stiber}
\bibfield{author}{\bibinfo{person}{Maia Stiber}, \bibinfo{person}{Russell Taylor}, {and} \bibinfo{person}{Chien-Ming Huang}.} \bibinfo{year}{2022}\natexlab{}.
\newblock \showarticletitle{Modeling Human Response to Robot Errors for Timely Error Detection}. In \bibinfo{booktitle}{\emph{2022 IEEE/RSJ International Conference on Intelligent Robots and Systems (IROS)}}. \bibinfo{publisher}{IEEE}, \bibinfo{address}{New York, NY, USA}, \bibinfo{pages}{676--683}.
\newblock
\urldef\tempurl%
\url{https://doi.org/10.1109/IROS47612.2022.9981726}
\showDOI{\tempurl}


\bibitem[Stiber et~al\mbox{.}(2023)]%
        {stiber2023using}
\bibfield{author}{\bibinfo{person}{Maia Stiber}, \bibinfo{person}{Russell~H Taylor}, {and} \bibinfo{person}{Chien-Ming Huang}.} \bibinfo{year}{2023}\natexlab{}.
\newblock \showarticletitle{On Using Social Signals to Enable Flexible Error-Aware HRI}.
\newblock  (\bibinfo{year}{2023}).
\newblock


\bibitem[Sun et~al\mbox{.}(2020)]%
        {sun2020eev}
\bibfield{author}{\bibinfo{person}{Jennifer~J Sun}, \bibinfo{person}{Ting Liu}, \bibinfo{person}{Alan~S Cowen}, \bibinfo{person}{Florian Schroff}, \bibinfo{person}{Hartwig Adam}, {and} \bibinfo{person}{Gautam Prasad}.} \bibinfo{year}{2020}\natexlab{}.
\newblock \showarticletitle{EEV: A large-scale dataset for studying evoked expressions from video}.
\newblock \bibinfo{journal}{\emph{arXiv preprint arXiv:2001.05488}} (\bibinfo{year}{2020}).
\newblock


\bibitem[Sutton and Barto(2018)]%
        {sutton2018reinforcement}
\bibfield{author}{\bibinfo{person}{RS Sutton} {and} \bibinfo{person}{AG Barto}.} \bibinfo{year}{2018}\natexlab{}.
\newblock \bibinfo{title}{Reinforcement Learning: An Introduction. Cambridge, MA, USA: A Bradford Book}.
\newblock
\newblock


\bibitem[Sutton and Barto(1998)]%
        {sutton1998reinforcement}
\bibfield{author}{\bibinfo{person}{Richard~S Sutton} {and} \bibinfo{person}{Andrew~G Barto}.} \bibinfo{year}{1998}\natexlab{}.
\newblock \showarticletitle{Reinforcement learning: an introduction MIT Press}.
\newblock \bibinfo{journal}{\emph{Cambridge, MA}}  \bibinfo{volume}{22447} (\bibinfo{year}{1998}).
\newblock


\bibitem[Synakowski et~al\mbox{.}(2021)]%
        {synakowski2021adding}
\bibfield{author}{\bibinfo{person}{Stuart Synakowski}, \bibinfo{person}{Qianli Feng}, {and} \bibinfo{person}{Aleix Martinez}.} \bibinfo{year}{2021}\natexlab{}.
\newblock \showarticletitle{Adding knowledge to unsupervised algorithms for the recognition of intent}.
\newblock \bibinfo{journal}{\emph{International Journal of Computer Vision}} \bibinfo{volume}{129}, \bibinfo{number}{4} (\bibinfo{year}{2021}), \bibinfo{pages}{942--959}.
\newblock


\bibitem[Takayama et~al\mbox{.}(2011)]%
        {takayama2011expressing}
\bibfield{author}{\bibinfo{person}{Leila Takayama}, \bibinfo{person}{Doug Dooley}, {and} \bibinfo{person}{Wendy Ju}.} \bibinfo{year}{2011}\natexlab{}.
\newblock \showarticletitle{Expressing thought: improving robot readability with animation principles}. In \bibinfo{booktitle}{\emph{Proceedings of the 6th international conference on Human-robot interaction}}. \bibinfo{publisher}{ACM}, \bibinfo{address}{New York, NY, USA}, \bibinfo{pages}{69--76}.
\newblock


\bibitem[Tatnall(2005)]%
        {tatnall2005actor}
\bibfield{author}{\bibinfo{person}{Arthur Tatnall}.} \bibinfo{year}{2005}\natexlab{}.
\newblock \showarticletitle{Actor-network theory in information systems research}.
\newblock In \bibinfo{booktitle}{\emph{Encyclopedia of Information Science and Technology, First Edition}}. \bibinfo{publisher}{IGI Global}, \bibinfo{pages}{42--46}.
\newblock


\bibitem[Tian and Oviatt(2021)]%
        {tian2021taxonomy}
\bibfield{author}{\bibinfo{person}{Leimin Tian} {and} \bibinfo{person}{Sharon Oviatt}.} \bibinfo{year}{2021}\natexlab{}.
\newblock \showarticletitle{A Taxonomy of Social Errors in Human-Robot Interaction}.
\newblock \bibinfo{journal}{\emph{J. Hum.-Robot Interact.}} \bibinfo{volume}{10}, \bibinfo{number}{2}, Article \bibinfo{articleno}{13} (\bibinfo{date}{feb} \bibinfo{year}{2021}), \bibinfo{numpages}{32}~pages.
\newblock
\urldef\tempurl%
\url{https://doi.org/10.1145/3439720}
\showDOI{\tempurl}


\bibitem[Trung et~al\mbox{.}(2017)]%
        {2017trung}
\bibfield{author}{\bibinfo{person}{Pauline Trung}, \bibinfo{person}{Manuel Giuliani}, \bibinfo{person}{Michael Miksch}, \bibinfo{person}{Gerald Stollnberger}, \bibinfo{person}{Susanne Stadler}, \bibinfo{person}{Nicole Mirnig}, {and} \bibinfo{person}{Manfred Tscheligi}.} \bibinfo{year}{2017}\natexlab{}.
\newblock \showarticletitle{Head and shoulders: Automatic error detection in human-robot interaction}.
\newblock \bibinfo{journal}{\emph{Proceedings of International Conference on Multimodal Interaction}} (\bibinfo{year}{2017}).
\newblock
\showISBNx{9781450355438}


\bibitem[Urakami and Seaborn(2023)]%
        {urakami2023nonverbal}
\bibfield{author}{\bibinfo{person}{Jacqueline Urakami} {and} \bibinfo{person}{Katie Seaborn}.} \bibinfo{year}{2023}\natexlab{}.
\newblock \showarticletitle{Nonverbal Cues in Human–Robot Interaction: A Communication Studies Perspective}.
\newblock \bibinfo{journal}{\emph{J. Hum.-Robot Interact.}} \bibinfo{volume}{12}, \bibinfo{number}{2}, Article \bibinfo{articleno}{22} (\bibinfo{date}{mar} \bibinfo{year}{2023}), \bibinfo{numpages}{21}~pages.
\newblock
\urldef\tempurl%
\url{https://doi.org/10.1145/3570169}
\showDOI{\tempurl}


\bibitem[van Waveren et~al\mbox{.}(2019)]%
        {2019sanne}
\bibfield{author}{\bibinfo{person}{Sanne van Waveren}, \bibinfo{person}{Elizabeth~J. Carter}, {and} \bibinfo{person}{Iolanda Leite}.} \bibinfo{year}{2019}\natexlab{}.
\newblock \showarticletitle{Take One For the Team: The Effects of Error Severity in Collaborative Tasks with Social Robots}. In \bibinfo{booktitle}{\emph{Proceedings of the 19th ACM International Conference on Intelligent Virtual Agents}} (Paris, France) \emph{(\bibinfo{series}{IVA '19})}. \bibinfo{publisher}{Association for Computing Machinery}, \bibinfo{address}{New York, NY, USA}, \bibinfo{pages}{151–158}.
\newblock
\showISBNx{9781450366724}
\urldef\tempurl%
\url{https://doi.org/10.1145/3308532.3329475}
\showDOI{\tempurl}


\bibitem[Vinanzi et~al\mbox{.}(2021)]%
        {vinanzi2021collaborative}
\bibfield{author}{\bibinfo{person}{Samuele Vinanzi}, \bibinfo{person}{Angelo Cangelosi}, {and} \bibinfo{person}{Christian Goerick}.} \bibinfo{year}{2021}\natexlab{}.
\newblock \showarticletitle{The collaborative mind: intention reading and trust in human-robot interaction}.
\newblock \bibinfo{journal}{\emph{Iscience}} \bibinfo{volume}{24}, \bibinfo{number}{2} (\bibinfo{year}{2021}), \bibinfo{pages}{102130}.
\newblock


\bibitem[Walker(2022)]%
        {walker_2022}
\bibfield{author}{\bibinfo{person}{Barnabas~James Walker}.} \bibinfo{year}{2022}\natexlab{}.
\newblock
\newblock
\urldef\tempurl%
\url{https://www.citationgecko.com}
\showURL{%
\tempurl}


\bibitem[Wallk{\"o}tter et~al\mbox{.}(2021)]%
        {wallkotter2021explainable}
\bibfield{author}{\bibinfo{person}{Sebastian Wallk{\"o}tter}, \bibinfo{person}{Silvia Tulli}, \bibinfo{person}{Ginevra Castellano}, \bibinfo{person}{Ana Paiva}, {and} \bibinfo{person}{Mohamed Chetouani}.} \bibinfo{year}{2021}\natexlab{}.
\newblock \showarticletitle{Explainable embodied agents through social cues: a review}.
\newblock \bibinfo{journal}{\emph{ACM Transactions on Human-Robot Interaction (THRI)}} \bibinfo{volume}{10}, \bibinfo{number}{3} (\bibinfo{year}{2021}), \bibinfo{pages}{1--24}.
\newblock


\bibitem[Wang and Loewen(2016)]%
        {wang2016nonverbal}
\bibfield{author}{\bibinfo{person}{Weiqing Wang} {and} \bibinfo{person}{Shawn Loewen}.} \bibinfo{year}{2016}\natexlab{}.
\newblock \showarticletitle{Nonverbal behavior and corrective feedback in nine ESL university-level classrooms}.
\newblock \bibinfo{journal}{\emph{Language Teaching Research}} \bibinfo{volume}{20}, \bibinfo{number}{4} (\bibinfo{year}{2016}), \bibinfo{pages}{459--478}.
\newblock


\bibitem[Wang et~al\mbox{.}(2016)]%
        {wang2016human}
\bibfield{author}{\bibinfo{person}{Wen-June Wang}, \bibinfo{person}{Jun-Wei Chang}, \bibinfo{person}{Shih-Fu Haung}, {and} \bibinfo{person}{Rong-Jyue Wang}.} \bibinfo{year}{2016}\natexlab{}.
\newblock \showarticletitle{Human posture recognition based on images captured by the kinect sensor}.
\newblock \bibinfo{journal}{\emph{International Journal of Advanced Robotic Systems}} \bibinfo{volume}{13}, \bibinfo{number}{2} (\bibinfo{year}{2016}), \bibinfo{pages}{54}.
\newblock


\bibitem[Wang et~al\mbox{.}(2022)]%
        {wang2022systematic}
\bibfield{author}{\bibinfo{person}{Yan Wang}, \bibinfo{person}{Wei Song}, \bibinfo{person}{Wei Tao}, \bibinfo{person}{Antonio Liotta}, \bibinfo{person}{Dawei Yang}, \bibinfo{person}{Xinlei Li}, \bibinfo{person}{Shuyong Gao}, \bibinfo{person}{Yixuan Sun}, \bibinfo{person}{Weifeng Ge}, \bibinfo{person}{Wei Zhang}, {et~al\mbox{.}}} \bibinfo{year}{2022}\natexlab{}.
\newblock \showarticletitle{A systematic review on affective computing: Emotion models, databases, and recent advances}.
\newblock \bibinfo{journal}{\emph{Information Fusion}} (\bibinfo{year}{2022}).
\newblock


\bibitem[Weber et~al\mbox{.}(2018)]%
        {weber2018shape}
\bibfield{author}{\bibinfo{person}{Klaus Weber}, \bibinfo{person}{Hannes Ritschel}, \bibinfo{person}{Ilhan Aslan}, \bibinfo{person}{Florian Lingenfelser}, {and} \bibinfo{person}{Elisabeth Andr{\'e}}.} \bibinfo{year}{2018}\natexlab{}.
\newblock \showarticletitle{How to shape the humor of a robot-social behavior adaptation based on reinforcement learning}. In \bibinfo{booktitle}{\emph{Proceedings of the 20th ACM international conference on multimodal interaction}}. \bibinfo{publisher}{ACM}, \bibinfo{address}{New York, NY, USA}, \bibinfo{pages}{154--162}.
\newblock


\bibitem[Xiao et~al\mbox{.}(2014)]%
        {xiao2014human}
\bibfield{author}{\bibinfo{person}{Yang Xiao}, \bibinfo{person}{Zhijun Zhang}, \bibinfo{person}{Aryel Beck}, \bibinfo{person}{Junsong Yuan}, {and} \bibinfo{person}{Daniel Thalmann}.} \bibinfo{year}{2014}\natexlab{}.
\newblock \showarticletitle{Human--robot interaction by understanding upper body gestures}.
\newblock \bibinfo{journal}{\emph{Presence}} \bibinfo{volume}{23}, \bibinfo{number}{2} (\bibinfo{year}{2014}), \bibinfo{pages}{133--154}.
\newblock


\bibitem[Xu et~al\mbox{.}(2022)]%
        {xu2022probabilistic}
\bibfield{author}{\bibinfo{person}{Jinglin Xu}, \bibinfo{person}{Guangyi Chen}, \bibinfo{person}{Nuoxing Zhou}, \bibinfo{person}{Wei-Shi Zheng}, {and} \bibinfo{person}{Jiwen Lu}.} \bibinfo{year}{2022}\natexlab{}.
\newblock \showarticletitle{Probabilistic Temporal Modeling for Unintentional Action Localization}.
\newblock \bibinfo{journal}{\emph{IEEE Transactions on Image Processing}}  \bibinfo{volume}{31} (\bibinfo{year}{2022}), \bibinfo{pages}{3081--3094}.
\newblock


\bibitem[Yasuda and Matsumoto(2013)]%
        {2013yasuda}
\bibfield{author}{\bibinfo{person}{Hiroyuki Yasuda} {and} \bibinfo{person}{Mitsuharu Matsumoto}.} \bibinfo{year}{2013}\natexlab{}.
\newblock \showarticletitle{Psychological impact on human when a robot makes mistakes}. In \bibinfo{booktitle}{\emph{Proceedings of the 2013 IEEE/SICE International Symposium on System Integration}}. \bibinfo{publisher}{IEEE}, \bibinfo{address}{New York, NY, USA}, \bibinfo{pages}{335--339}.
\newblock
\urldef\tempurl%
\url{https://doi.org/10.1109/SII.2013.6776612}
\showDOI{\tempurl}


\bibitem[Zeng et~al\mbox{.}(2004)]%
        {zeng2004bimodal}
\bibfield{author}{\bibinfo{person}{Zhihong Zeng}, \bibinfo{person}{Jilin Tu}, \bibinfo{person}{Ming Liu}, \bibinfo{person}{Tong Zhang}, \bibinfo{person}{Nicholas Rizzolo}, \bibinfo{person}{Zhenqiu Zhang}, \bibinfo{person}{Thomas~S Huang}, \bibinfo{person}{Dan Roth}, {and} \bibinfo{person}{Stephen Levinson}.} \bibinfo{year}{2004}\natexlab{}.
\newblock \showarticletitle{Bimodal HCI-related affect recognition}. In \bibinfo{booktitle}{\emph{Proceedings of the 6th international conference on Multimodal interfaces}}. \bibinfo{pages}{137--143}.
\newblock


\bibitem[Zhang et~al\mbox{.}(2021)]%
        {zhang2021real}
\bibfield{author}{\bibinfo{person}{Ke Zhang}, \bibinfo{person}{Yuanqing Li}, \bibinfo{person}{Jingyu Wang}, \bibinfo{person}{Erik Cambria}, {and} \bibinfo{person}{Xuelong Li}.} \bibinfo{year}{2021}\natexlab{}.
\newblock \showarticletitle{Real-time video emotion recognition based on reinforcement learning and domain knowledge}.
\newblock \bibinfo{journal}{\emph{IEEE Transactions on Circuits and Systems for Video Technology}} \bibinfo{volume}{32}, \bibinfo{number}{3} (\bibinfo{year}{2021}), \bibinfo{pages}{1034--1047}.
\newblock


\bibitem[Zhang et~al\mbox{.}(2023)]%
        {zhang2023self}
\bibfield{author}{\bibinfo{person}{Qiping Zhang}, \bibinfo{person}{Austin Narcomey}, \bibinfo{person}{Kate Candon}, {and} \bibinfo{person}{Marynel V\'{a}zquez}.} \bibinfo{year}{2023}\natexlab{}.
\newblock \showarticletitle{Self-Annotation Methods for Aligning Implicit and Explicit Human Feedback in Human-Robot Interaction}. In \bibinfo{booktitle}{\emph{Proceedings of the 2023 ACM/IEEE International Conference on Human-Robot Interaction}} \emph{(\bibinfo{series}{HRI '23})}. \bibinfo{publisher}{Association for Computing Machinery}, \bibinfo{address}{New York, NY, USA}, \bibinfo{pages}{398–407}.
\newblock
\showISBNx{9781450399647}
\urldef\tempurl%
\url{https://doi.org/10.1145/3568162.3576986}
\showDOI{\tempurl}


\bibitem[Zhang et~al\mbox{.}(2014)]%
        {zhang2014bp4d}
\bibfield{author}{\bibinfo{person}{Xing Zhang}, \bibinfo{person}{Lijun Yin}, \bibinfo{person}{Jeffrey~F Cohn}, \bibinfo{person}{Shaun Canavan}, \bibinfo{person}{Michael Reale}, \bibinfo{person}{Andy Horowitz}, \bibinfo{person}{Peng Liu}, {and} \bibinfo{person}{Jeffrey~M Girard}.} \bibinfo{year}{2014}\natexlab{}.
\newblock \showarticletitle{Bp4d-spontaneous: a high-resolution spontaneous 3d dynamic facial expression database}.
\newblock \bibinfo{journal}{\emph{Image and Vision Computing}} \bibinfo{volume}{32}, \bibinfo{number}{10} (\bibinfo{year}{2014}), \bibinfo{pages}{692--706}.
\newblock


\bibitem[Zhao et~al\mbox{.}(2011)]%
        {zhao2011facial}
\bibfield{author}{\bibinfo{person}{Guoying Zhao}, \bibinfo{person}{Xiaohua Huang}, \bibinfo{person}{Matti Taini}, \bibinfo{person}{Stan~Z Li}, {and} \bibinfo{person}{Matti Pietik{\"a}Inen}.} \bibinfo{year}{2011}\natexlab{}.
\newblock \showarticletitle{Facial expression recognition from near-infrared videos}.
\newblock \bibinfo{journal}{\emph{Image and vision computing}} \bibinfo{volume}{29}, \bibinfo{number}{9} (\bibinfo{year}{2011}), \bibinfo{pages}{607--619}.
\newblock


\bibitem[Zhou et~al\mbox{.}(2021)]%
        {zhou2021temporal}
\bibfield{author}{\bibinfo{person}{Nuoxing Zhou}, \bibinfo{person}{Guangyi Chen}, \bibinfo{person}{Jinglin Xu}, \bibinfo{person}{Wei-Shi Zheng}, {and} \bibinfo{person}{Jiwen Lu}.} \bibinfo{year}{2021}\natexlab{}.
\newblock \showarticletitle{Temporal label aggregation for unintentional action localization}. In \bibinfo{booktitle}{\emph{2021 IEEE International Conference on Multimedia and Expo (ICME)}}. IEEE, \bibinfo{publisher}{IEEE}, \bibinfo{address}{New York, NY, USA}, \bibinfo{pages}{1--7}.
\newblock


\end{thebibliography}

\end{document}